\documentclass[12pt]{article}

\usepackage{amsmath,amssymb, bbm}
\usepackage{upgreek}
\usepackage{helvet}

\usepackage{multirow}
\usepackage{graphicx}
\usepackage{lineno}
\usepackage{xr}
\usepackage[left=0.75in,top=0.75in,right=0.75in, bottom=0.75in]{geometry}
\usepackage{multirow}%
\usepackage{amsmath,amssymb,amsfonts}%
\usepackage{amsthm}%
\usepackage{mathrsfs}%
\usepackage[title]{appendix}%
\usepackage{xcolor}%
\usepackage{textcomp}%
\usepackage{manyfoot}%
\usepackage{booktabs}%
\usepackage{algorithm}%
\usepackage{algorithmicx}%
\usepackage{algpseudocode}%
\usepackage{listings}%
\usepackage{array}
\usepackage{cite}
\usepackage{color}
\usepackage{setspace}
\usepackage{booktabs}
\usepackage{amssymb, bm, booktabs, hyperref}
\usepackage{upgreek}
\usepackage{todonotes}
\usepackage{adjustbox}
\makeatletter
\renewcommand{\@biblabel}[1]{\quad#1.}
\makeatother

\newcommand{\E}{\mathbb{E}}

\usepackage{url}
\usepackage{multirow}
\usepackage{setspace}

\usepackage[utf8]{inputenc} 
\usepackage[T1]{fontenc}    
\usepackage{url}            
\usepackage{amsfonts}       
\usepackage{nicefrac}       
\usepackage{microtype}      

\definecolor{myblue}{RGB}{36,86,166}
\definecolor{myred}{RGB}{147,31,37}
\definecolor{myblue2}{RGB}{140,180,200}
\definecolor{myred2}{RGB}{213,23,40}

\definecolor{DarkRed}{rgb}{0.5,0.1,0.1}

\usepackage[caption=true]{subfig}

\usepackage{hyperref}

\usepackage{authblk}

\usepackage{abstract}

\usepackage{multibib}

\title{scE$^2$TM improves single-cell embedding interpretability and reveals cellular perturbation signatures}
\author[1,2]{Hegang Chen}
\author[1]{Yuyin Lu}
\author[3]{Yifan Zhao}
\author[1]{Zhiming Dai}
\author[4]{Fu Lee Wang}
\author[5]{Qing Li}
\author[1,*]{Yanghui Rao}
\author[2,*]{Yue Li}
\affil[1]{School of Computer Science and Engineering, Sun Yat-sen University, Guangzhou, China.}
\affil[2]{School of Computer Science, McGill University, Montreal, Canada}
\affil[3]{Department of Biomedical Informatics, Harvard Medical School, Boston, USA.}
\affil[4]{School of Science and Technology, Hong Kong Metropolitan
University, Hong Kong, China.}
\affil[5]{Department of Computing, The Hong Kong Polytechnic University, Hong Kong, China.}
\affil[*]{Correspondence to raoyangh@mail.sysu.edu.cn, yueli@cs.mcgill.ca}

\date{}



\begin{document}

\maketitle

\setlength{\parindent}{1em}

\begin{abstract}
\begin{center}
	\section*{Abstract}
\end{center}	
Single-cell RNA sequencing technologies have revolutionized our understanding of cellular heterogeneity, yet computational methods often struggle to balance performance with biological interpretability. 
Embedded topic models have been widely used for interpretable single-cell embedding learning. 
However, these models suffer from the potential problem of interpretation collapse, where topics semantically collapse towards each other, resulting in redundant topics and incomplete capture of biological variation. Furthermore, the rise of single-cell foundation models creates opportunities to harness external biological knowledge for guiding model embeddings. Here, we present scE$^2$TM, an external knowledge-guided embedded topic model that provides a high-quality cell embedding and interpretation for scRNA-seq analysis. Through embedding clustering regularization method, each topic is constrained to be the center of a separately aggregated gene cluster, enabling it to capture unique biological information.
Across 20 scRNA-seq datasets, scE$^2$TM  achieves superior clustering performance compared with seven state-of-the-art methods.
A comprehensive interpretability benchmark further shows that scE$^2$TM-learned topics exhibit higher diversity and stronger consistency with underlying biological pathways.
Modeling interferon-stimulated PBMCs, scE$^2$TM simulates topic perturbations that drive control cells toward stimulated-like transcriptional states, faithfully mirroring experimental interferon responses.
In melanoma, scE$^2$TM identifies malignant-specific topics and extrapolates them to unseen patient data, revealing gene programs associated with patient survival.
\end{abstract}

\section{Introduction}\label{sec1}
Advances in single-cell sequencing technologies have revolutionized our ability to study biological systems with unprecedented resolution \cite{kolodziejczyk2015technology,hwang2018single}. A central computational task in analyzing single-cell RNA sequencing (scRNA-seq) data is unsupervised clustering, which enables \textit{de novo} cell-type identification and has revealed previously uncharacterized cellular states \cite{tian2021model}. In recent years, deep learning methods have attracted significant attention in the single-cell community due to their remarkable ability to process sparse, noisy, and high-dimensional data \cite{ma2022deep}. However, the inherent complexity of multi-layer nonlinear architectures, activation functions, and large parameter spaces often makes these deep learning models difficult to understand and verify \cite{rudin2019stop}. 
This lack of interpretability poses a significant challenge to verifying whether proposed models can capture actual biological mechanisms and to translating computational findings into biological insights.
Interpretable deep learning has become a promising approach to address challenges in analyzing biological data. A model is deemed interpretable when its learned parameters can directly link input features with latent variables or target results \cite{zhao2021learning}, allowing researchers to gain biological insights without needing extensive \textit{post hoc} analyses \cite{klauschen2024toward,tseng2020fourier,tao2022interpretable}. Recently, topic modeling has been successfully applied in single-cell genomics \cite{zhao2021learning,swapna2023gtm,fan2024gfetm}.
Originally developed for mining textual data \cite{chen2023nonlinear,lu2024self,chen2025supervised}, topic modeling’s appeal lies in its ability to efficiently handle high-dimensional sparse data while providing interpretability by associating input features with latent factors or outcomes through discovered topics. In the context of single-cell omics analysis, a series of studies have integrated deep learning-based embedding techniques with topic modeling, leading to the development of various single-cell embedded topic models \cite{lynch2022mira,zhang2023unraveling,subedi2023single,chen2024comprehensive,kazwini2024share,fan2024gfetm,zhong2024interpretable}. scETM \cite{zhao2021learning} adapted the embedded topic modeling \cite{dieng2020topic}, utilizing interpretable linear decoders to learn highly interpretable gene and topic embeddings. Building upon this, d-scIGM \cite{chen2024comprehensive} enhances the model's capacity to capture latent data signals by employing deeper neural network architectures, thereby better modeling the dependencies between topics and genes. SPRUCE \cite{subedi2023single} combines a topic modeling approach with cell-cell interactions to simulate informational crosstalk between cell-cells while identifying cell types.

Despite this progress, existing embedded topic models tend to suffer from interpretation collapse, where the optimization disproportionately weights frequent elements during trainnig, causing topic embeddings to converge and lose diversity  \cite{zhaoneural,wu2023effective}. scRNA-seq data mirrors in an important way: just as text has a long-tailed word distribution, scRNA-seq data is usually dominated by highly expressed housekeeping genes that lack cell-type-specificity. This creates conditions where interpretation collapse can occur.  Interpretation collapse generally manifests in two ways. First, models over-emphasize highly expressed genes, leading to redundant identification of common gene programs while missing diverse biological signals in rare cell types. Second, the learned topics may not always align well with actual cell types. In other words, if the topic assignments do not align well with established cell-type annotations, it suggests that the models are failing to capture a biologically meaningful representation of cellular identity. Current interpretability assessments for most single-cell embedding topic models rely on qualitative analysis of model-derived topics \cite{chen2024applying,wagle2024interpretable}. Such an approach carries inherent risks of bias and subjective interpretation, making it difficult to systematically determine whether model optimization preserves interpretation quality.
Meanwhile, recent advances in single-cell foundation models trained on massive databases \cite{cui2024scgpt,hao2024large} have demonstrated robust performance across numerous downstream tasks, offering a promising source of diverse knowledge as supervisory signals for embedding and clustering. The rich knowledge of single-cell foundation models can effectively guide embedding and clustering by providing external supervision beyond the constraints of individual datasets.

\begin{figure*}
    \centering  
    \includegraphics[width=1\linewidth]{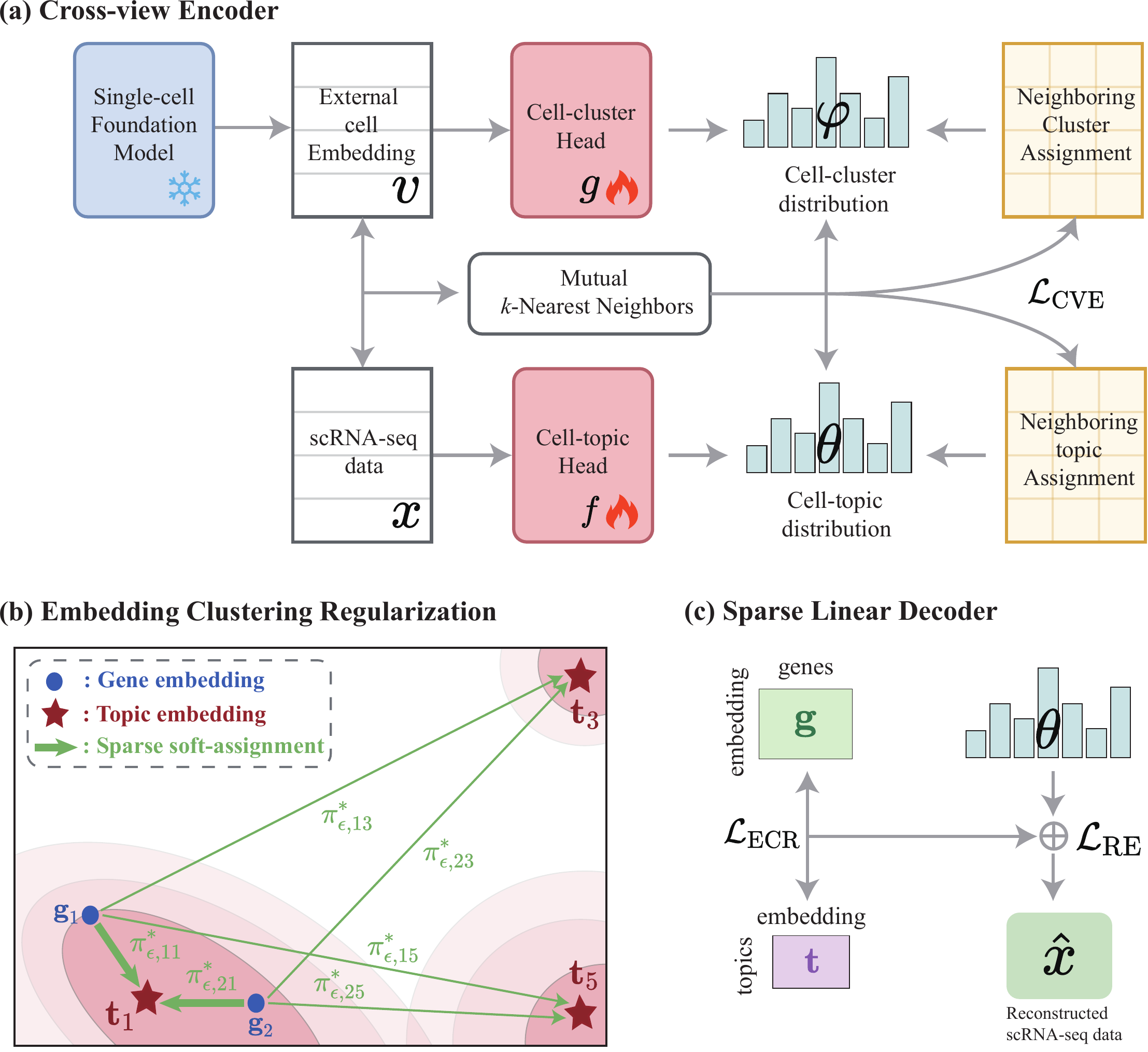}
    \caption{Schematic overview of scE$^2$TM. (a) Cross-view encoder. This encoder integrates single-cell expression data with embeddings extracted from a single-cell foundation model. To combine these two perspectives, cluster and topic heads are trained based on the mutual neighborhood information by encouraging consistent clustering assignments of mutual nearest neighbors of the corresponding cells. (b)  Embedding clustering regularization module. ECR clusters gene embeddings $\mathbf{g}_{j}$ ($\textcolor{myblue}{\bullet}$) as samples and topic embeddings $\mathbf{t}_{k}$ ($\textcolor{myred}{\star}$) as centers with soft-assignment $\pi_{\epsilon, j k}^{*}$. For instance, ECR pushes $\mathbf{g}_{1}$ and $\mathbf{g}_{2}$ close to $\mathbf{t}_{1}$ and away from $\mathbf{t}_{3}$ and $\mathbf{t}_{5}$. (c) Sparse linear decoder. Decoder learns topic embeddings and gene embeddings as well as sparse topic-gene dependencies during reconstruction, thereby ensuring model interpretability.}
    \label{fig:model}
\end{figure*}

We present \textbf{scE$^2$TM} (\textbf{s}ingle-\textbf{c}ell \textbf{E}xternal knowledge-guided \textbf{E}mbedding clustering regularization \textbf{T}opic \textbf{M}odel), a deep generative model for large-scale single-cell transcriptomic analysis (Fig. \ref{fig:model}). scE$^2$TM introduces two key innovations to overcome interpretation collapse. First, \textbf{E}mbedding \textbf{C}lustering \textbf{R}egularization (\textbf{ECR}) \cite{wu2023effective} encourages topic diversity by regularizing topic embeddings as clustering centers and gene embeddings into clustering samples, with cluster soft assignments modeled through Optimal Transport (OT) and enhances representational learning of distinct biological processes. 
Second, scE$^2$TM integrates external knowledge from single-cell foundation models through a \textbf{C}ross-\textbf{v}iew \textbf{E}ncoder (\textbf{CVE}), shifting from internal guidance paradigms to scalable external knowledge integration. 
To systematically evaluate scE$^2$TM, we employed ten quantitative interpretability metrics that, combined with clustering metrics, enable rigorous assessment of topic model performance.
Benchmarking across 20 datasets revealed that scE$^2$TM mitigates interpretation collapse while existing methods suffer. Critically, clustering performance and interpretability are mostly uncorrelated—demonstrating that previous models sacrifice bioloical insights while optimizing clustering accuracy. scE$^2$TM achieved state-of-the-art performance in both aspects. We then showcase real-world applications of scE$^2$TM in three case studies.
In a human pancreas scRNA-seq dataset, scE$^2$TM captured insulin biosynthesis pathways and programs reflecting key cellular identities. 
Topic-based perturbations in interferon (IFN)-stimulated PBMCs successfully transformed cellular states,
validating capacity to simulate cellular responses to exogenous stimuli. In a melanoma scRNA-seq dataset, scE$^2$TM-derived topics generalized to TCGA cohorts with topic activity significantly correlating with patient survival, demonstrating clinical relevance.
Collectively, these results establish scE$^2$TM's as a framework for extracting mechanistically insightful and clinically actionable patterns from single-cell transcriptomics.

\section{Results}\label{sec2}
\subsection{Overview of scE$^2$TM and interpretability evaluation benchmarks}
We present scE$^2$TM, a single-cell embedded topic model designed for modeling scRNA-seq data, together with comprehensive framework for quantitatively evaluating model interpretability. While existing single-cell topic models have provided valuable foundations for the field, three opportunities remain for advancing the state-of-the-art: integrating external biological knowledge from foundation models to enhance performance, addressing topic interpretation collapse where models may generate overlapping topics, and establishing systematic methods for quantifying interpretation quality.

To address these challenges, scE$^2$TM provides three innovations. First, we introduce a Cross-view Encoder (CVE) which systematically transfers knowledge from single-cell foundation models into the embedding space. The CVE aligns the internal transcriptomic features (scRNA-seq transcriptome, $\mathbf{x}$ ) with external embedding ($\mathbf{v}$)  derived from a foundation model through cross-view mutual distillation which integrates different viewpoints through mutual distillation of neighborhood information. By enforcing consistent topic and cluster assignments between corresponding cells and their neighbors across views (Fig. \ref{fig:model}a), CVE effectively incorporates contextual biological knowledge learned from millions of cells into the model's representation learning process. Second, we implement Embedding Clustering Regularization (ECR) \cite{wu2023effective} to prevent interpretation collapse and enhance topic diversity. ECR solves an Optimal Transport (OT) problem that establishes soft cluster assignments by linking topic embeddings with the gene embeddings (Fig. \ref{fig:model}b). This formulation compels each topic embedding to be unique cluster center aggregating unique gene embeddings, ensuring topics remain distinct and span diverse semantic spaces. Cell representations are then reconstructed using the distribution of learned topics and the optimized topic-gene dependencies between topics and genes facilitated by ECR (Fig. \ref{fig:model}c).
Third, we conduct a systematic interpretability assessment framework comprising ten quantitative metrics to evaluate topics from multiple perspectives, such as alignment with cell types, intrinsic coherence and diversity, and the capacity to capture biological pathways.

\subsection{Robust and ablation analysis of scE$^2$TM}
We systematically assessed the robustness of scE$^2$TM by conducting extensive sensitivity analyses on key hyperparameters based on clustering and interpretability metrics (Supplementary Fig. \ref{supf6}; Section \ref{sec:cluster-metric}, \ref{sec:interpre-metric}). To integrate different viewpoints, CVE employs mutual distillation of neighborhood information. We observed stable clustering and interpretability performance across various numbers of nearest neighbors, suggesting the model's robustness to the choice of neighborhood size (Supplementary Fig. \ref{supf6}a). Subsequently, we analyzed the impact of the regularization term weights in ECR on performance. While increasing the weights induces some changes in the performance of topics and the related pathways for topic mining, the overall stability is maintained (Supplementary Fig. \ref{supf6}b). Finally, we visualized how $\lambda$ (i.e., regularization weight used in ECR) influences the topic-gene geometry on the human pancreas data via UMAP \cite{wang2016single} (Supplementary Fig. \ref{supf6}c). As $\lambda$ increases, the topic distribution becomes more discrete, with genes confined to the neighborhood of the topic.

To further investigate the contribution of each component in scE$^2$TM, we performed ablation studies across four scenarios: (i) omitting the embedding clustering regularization module (without ECR), (ii) excluding the cross-view encoder (without CVE), (iii) substituting the cross-view distillation strategy with embedded connectivity (with EC), and (iv) utilizing cell embeddings derived from a foundational model scGPT for clustering. Supplementary Table \ref{supt2} presents the average metrics for these four ablation scenarios of scE$^2$TM. The ECR ablation notably decreases the diversity of interpretations, indicating its essential role in preventing interpretation collapse. The ECR module achieves this by minimizing the repetition of key genes across topics, ensuring geometric compactness between topic embeddings and their related gene embeddings. Over-representation analysis (ORA), concentrating on the top genes, mirrors the topic metrics such as Topic Coherence (TC) versus ORA$_N$ and Topic Diversity (TD) versus ORA$_U$. Unlike ORA, GSEA shows different results when the ECR module is removed. Without the ECR, the distance between a topic and its top gene embedding is not constrained, increasing the repetition rate among top genes and reducing TD and ORA$_U$. Conversely, GSEA, which considers all genes, reveals that without ECR, genes are more evenly distributed in the embedding space. This means that even though the repetition rate of top genes in each topic rises, they still represent only a small part of the entire gene set, leading to contrasting outcomes. This difference underscores how various interpretability metrics capture different facets of biological coherence.

Eliminating CVE significantly worsens clustering performance, with ARI decreasing by 7.2\% and NMI by 4.8\%, whereas straightforward knowledge incorporation (using EC) offers only slight improvements. These findings validate the necessity of the proposed cross-view distillation framework for effectively integrating external knowledge with scRNA-seq data. Furthermore, the scGPT foundation model shows inferior performance in zero-shot scenarios, with ARI dropping by 25.0\% and NMI by 19.8\% compared to scE$^2$TM, underscoring the importance of adapting to specific tasks \cite{kedzierska2025zero}. Together, ablation studies reveal that scE$^2$TM consistently performs well across crucial hyperparameters and that both its ECR and CVE are crucial elements, jointly preventing interpretation failures and successfully merging knowledge from foundation models to achieve enhanced clustering and interpretability.

\begin{figure*}[t!]
    \centering
    \includegraphics[width=1\linewidth]{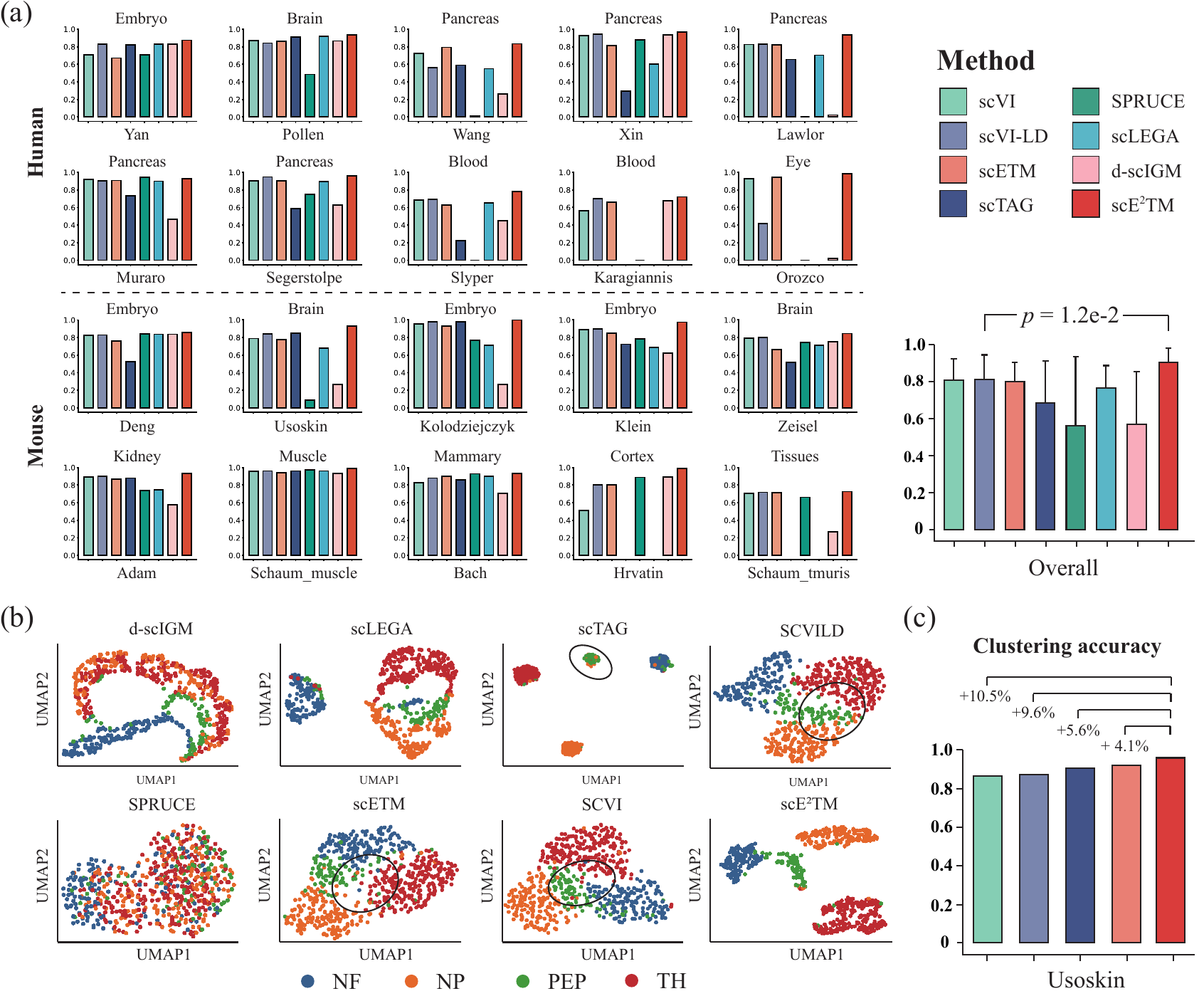}
    \caption{Cell-type clustering benchmark. (a) Cell type clustering performance evaluation in terms of Adjusted Random Index (ARI). The 20 panels on the left show the ARI values for the 8 benchmark methods on 20 datasets. The panel on the right displays the average ARI values and standard deviations. The statistical significance between scE$^2$M and the second best method was tested by pairwise Mann-Whitney U-test. Results for scTAG and scLEGA on some of the large scRNA-seq datasets (Karagiannis, Orozco, Hrvatin, and Schaum\_tmuris) were not shown because of their limited scalability. (b) UMAP visualization on the Usoskin dataset. UMAP was performed on the embeddings from each benchmark method on the Usoskin dataset. Cell types include peptidergic nociceptors (PEP), non-peptidergic nociceptors (NP), neurofilament (NF), and tyrosine hydroxylase (TH). (c) Cell-type clustering accuracy. The percentages of accuracy improvement achieved by scE$^2$TM relative to the baselines are labeled on the corresponding bar plots.}
    \label{figure2}
\end{figure*}

\subsection{scE$^2$TM provides high-quality cell embedding}
We evaluated scE$^2$TM against seven leading single-cell methods across 20 scRNA-seq datasets that span multiple species and diverse single-cell profiling technologies, with dataset sizes ranging from a few hundred to over 100,000 cells. scE$^2$TM substantially outperforms all state-of-the-art methods, achieving an overall Adjusted Rand Index (ARI) gain of 11.3\% ($p$-value = 1.2e-2) and Normalized Mutual Information (NMI) gain of 8.7\% ($p$-value = 2.1e-2) over the best-performing competitor, scVI-LD (Fig. \ref{figure2}a; Supplementary Fig. \ref{supf1}). Methods such as SPRUCE and d-scIGM show pronounced performance variability across datasets (Fig. \ref{figure2}a, error bars), indicating sensitivity to dataset-specific characteristics. In contrast, scE$^2$TM demonstrates robustness and consistent performance, with standard deviations of 0.082 and 0.066 for ARI and NMI, respectively. Notably, scE$^2$TM surpasses the cell network-integrated method, scLEGA, by 18.1\% ($p$-value = 5.2e-4) in ARI and 14.1\% ($p$-value = 4.6e-3) in NMI. This advantage stems from scE$^2$TM's external knowledge-guided embedding strategy and CVE framework, which more effectively leverages biological prior knowledge compared to scLEGA's dependence on internally built cell networks.
We also assessed the computational efficiency across three datasets of different sizes (Supplementary Table \ref{supt1}). The results indicate that scTAG and scLEGA based on cellular networks achieve fast runtimes but prohibitively high GPU memory usage.
Moreover, scVI and scVI-LD show more efficient speed and resource utilization, but both models exhibit poor interpretability. In contrast, scE$^2$TM achieves optimal clustering and interpretability while maintaining competitive computational efficiency.

We visualized cell embeddings from scE$^2$TM and competing methods on the Usoskin dataset \cite{usoskin2015unbiased}, which has well-defined cell types. scE$^2$TM's embeddings effectively capture biological variation across distinct conditions, whereas methods like scETM, scVI, and scVI-LD fail to distinguish overlapping populations (Fig. \ref{figure2}b,  black circles). We further quantified per-cell-type clustering accuracy when the predicted cluster number matches the true cell type count. scE$^2$TM achieves the highest mean accuracy, further validating its embedding quality (Fig. \ref{figure2}c).
Together, scE$^2$TM delivers robust, high-quality cell embeddings that consistently capture biologically meaningful structures across diverse datasets, outperforming existing state-of-the-art methods.


\begin{figure}[b!]
    \centering
    \includegraphics[width=\linewidth]{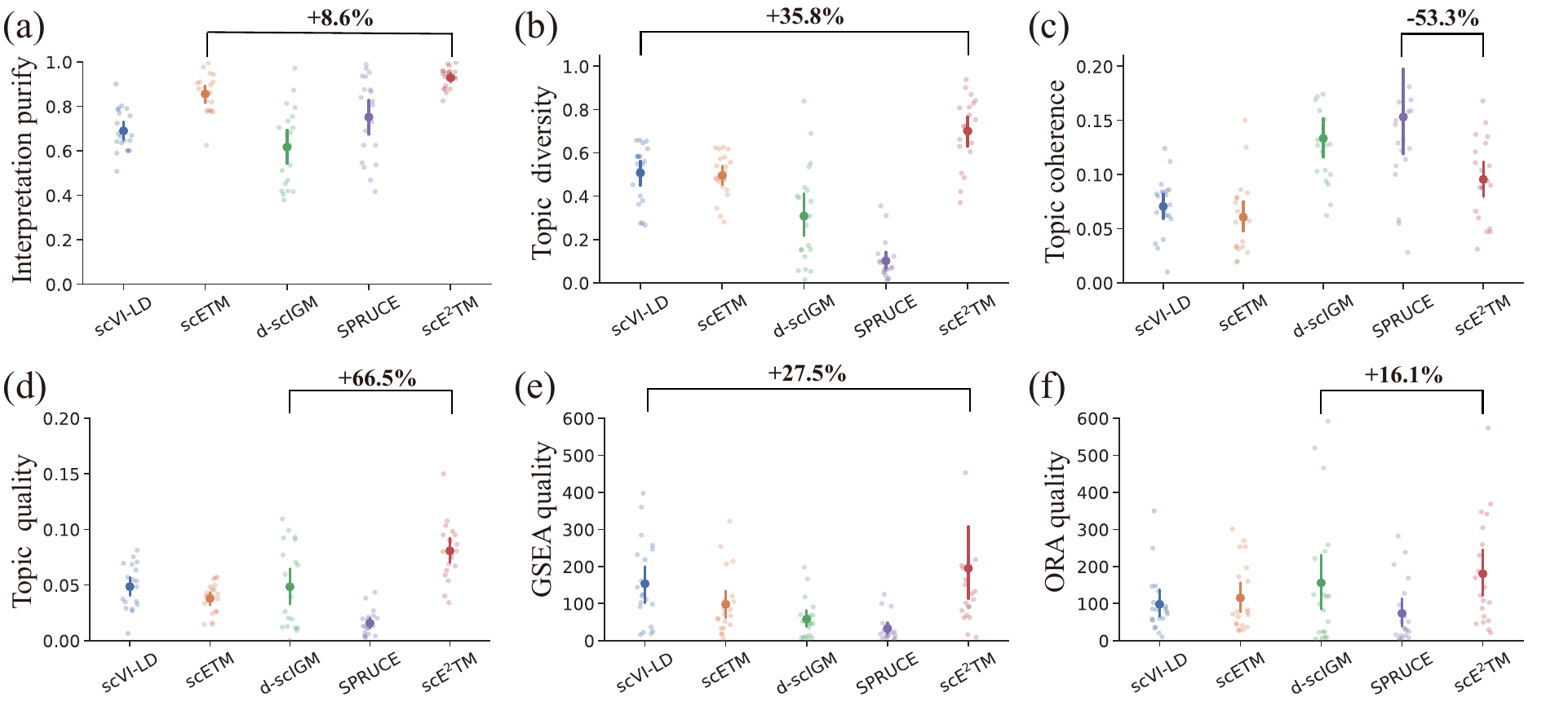}
    \caption{
    Interpretability metrics comparison. Quantitative assessment of interpretability using six metrics including Interpretation Purity (IP), Topic Diversity (TD), Topic Coherence (TC), Topic Quality (TQ), GSEA quality (GSEA$_Q$), and ORA quality (ORA$_Q$) across 20 scRNA-seq datasets. The percentage gain of scE$^2$TM over the second best method was marked on each panel.
    }
    \label{figure4}
\end{figure}

\begin{figure}[b!]
    \centering
    \includegraphics[width=\linewidth]{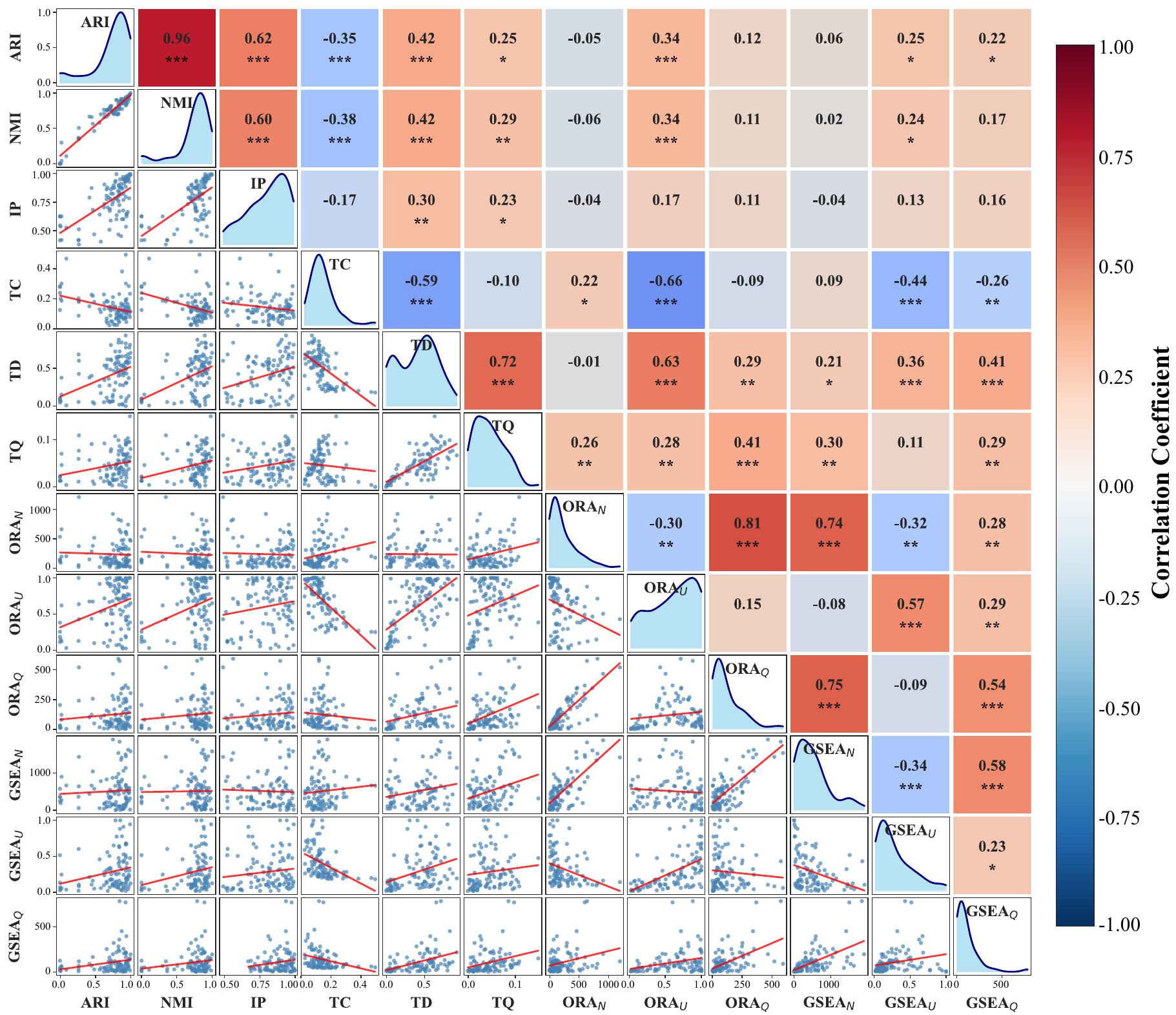}
    \caption{
   Correlation between interpretability and clustering metrics. Pairwise scatter plots in the lower triangle illustrate the relationships between each pair of metrics. Each point corresponds to the metric scores of one model-dataset pair, covering five single-cell embedded topic models applied to twenty datasets. The kernel density  along the diagonal estimates their marginal distributions. Pearson correlation coefficients ($r$) and their statistical significance are summarized in the accompanying heatmap in the upper triangle. Each cell reports the $r$ value together with its significance level ($p$ < 0.05: $*$, $p$ < 0.01: $**$, $p$ < 0.001: $***$).
    }
    \label{figure41}
\end{figure}

\subsection{Quantitative benchmark of model interpretability}
\label{sec:eval-interpre}

Recent studies \cite{bravo2023scenic+,zhou2023single,wagle2024interpretable} demonstrated the inherent risks of bias and cognitive limitations when relying primarily on qualitative methods to assess interpretability, underscoring the need for quantifying the biological interpretability.
We employed ten quantitative metrics (Section \ref{sec:interpre-metric}) to comprehensively evaluate the interpretability of single-cell embedded topic models:
Interpretation Purity (IP), Topic Diversity (TD), Topic Coherence (TC), Topic Quality (TQ), Number of pathways enriched by topic ORA (ORA$_N$), Uniqueness of pathways enriched by topic ORA (ORA$_U$), ORA Quality (ORA$_Q$), Number of pathways enriched by topic GSEA (GSEA$_N$), Uniqueness of pathways enriched by topic GSEA (GSEA$_U$), and GSEA Quality (GSEA$_Q$). scE$^2$TM achieves the best interpretable performance overall, substantially outperforming the second best method in terms of IP (8.6\% performance gain), TQ (66.5\% performance gain), GSEA$_Q$ (27.5\% performance gain) and ORA$_Q$ (16.1\% performance gain) (Fig. \ref{figure4}; Supplementary Fig. \ref{supf2}). scVI-LD and d-scIGM exhibit low interpretation purity (Fig. \ref{figure4}a), even compared to SPRUCE which has the lowest topic quality (Fig. \ref{figure4}d), indicating that their interpretations do not reflect the cell type information well and may not accurately capture real biological signals.
d-scIGM and SPRUCE showed low topic diversity (Fig. \ref{figure4}b), with values close to 0.1 in some datasets, while their topic coherence scores are relatively high (Fig. \ref{figure4}c). This indicates that these methods focus on only a few coherent topics and cannot comprehensively represent single-cell data. Quantitative analyses of GSEA and ORA output reveal that d-scIGM- and SPRUCE-derived topics are enriched for a large number of repetitive pathways (Supplementary Fig. \ref{supf2}). These results reveal that existing single-cell embedded topic models suffer from interpretation collapse—characterized by low topic diversity, poor cell-type alignment, and repetitive pathway enrichment—which scE$^2$TM effectively mitigates through superior performance across all interpretability metrics.

To understand the relationships among our interpretability metrics and their associations with existing clustering benchmarks, we analyzed Pearson's correlation between all evaluation metrics (Fig. \ref{figure41}). 
Interpretation purity strongly correlates with clustering metrics (ARI and NMI; $r > 0.6$, $p$-value $<$ 0.001) . This is expected as both metrics assess cell type coherence. However, topic quality does not correlate with clustering scores. For example, although scVI-LD achieved the second-best topic quality and clustering performance, it ranked second lowest in interpretation purity (Fig. \ref{figure2}a  and Fig. \ref{figure4}a,d).
Moreover, TQ and topic enrichment (i.e., ORA$_Q$, and GSEA$_Q$) show weak association with the clustering metrics, with all correlations falling below 0.3.
This indicates that high clustering performance does not imply biologically meaningful interpretations.
Topic quality is positively correlated with the quality of the topic enrichment analyses (ORA$_Q$ and GSEA$_Q$). Consistent with this observation, ORA and GSEA reveal similar trends. 
Lastly, we found negative correlation between topic coherence and topic diversity as well as between the number of pathways (ORA$_N$ / GSEA$_N$) and uniqueness of topic enrichment (ORA$_U$ / GSEA$_U$), which is consistent with previous work \cite{lin-etal-2024-hierarchical}. 
Overall, these quantitative results demonstrate scE$^2$TM's superior interpretability across multiple complementary metrics and reveal a clear discrepancy between clustering performance and biologically meaningful interpretability.

\begin{figure*}[t!]
    \centering  
    \includegraphics[width=1\linewidth]{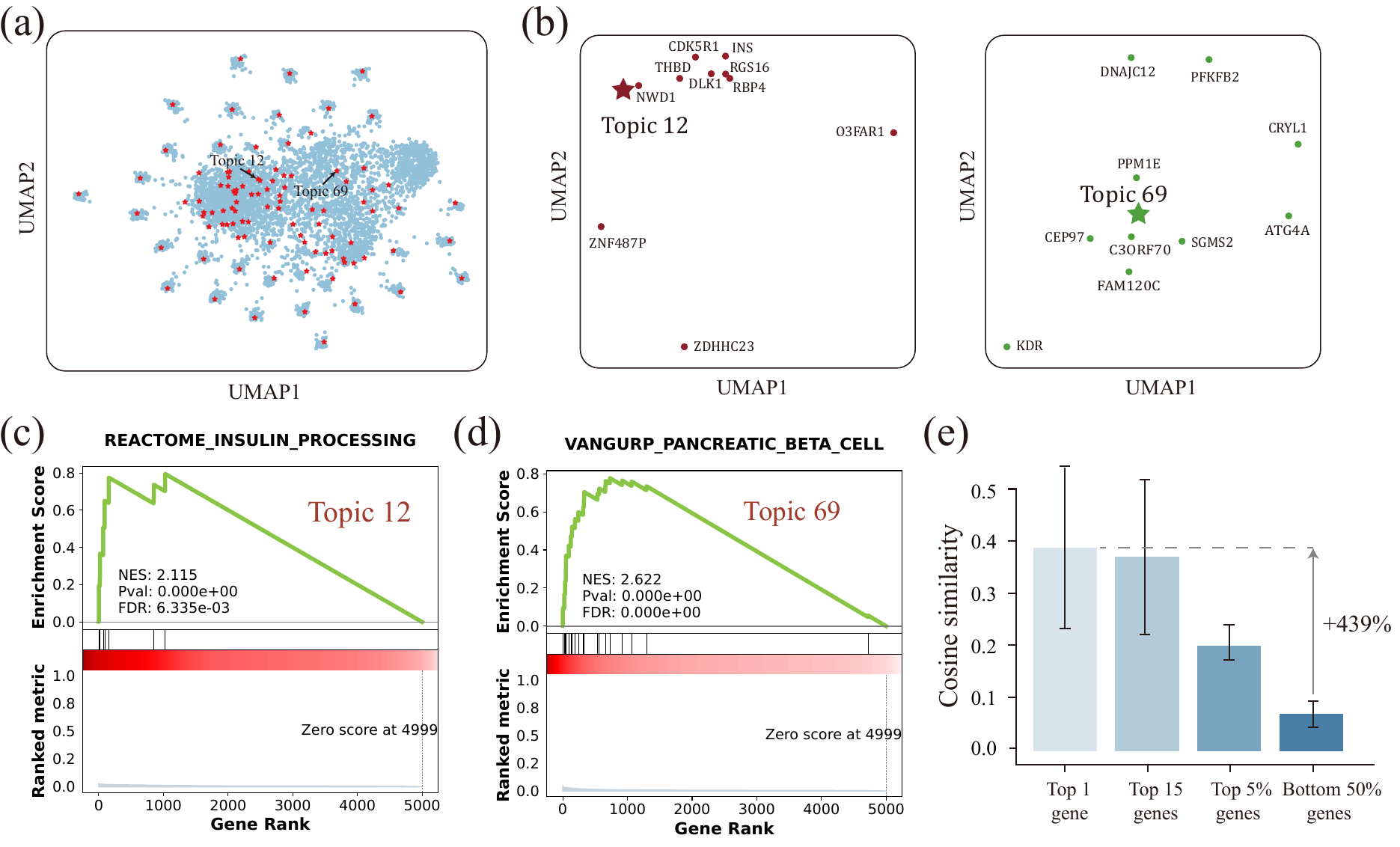}
    \caption{Analysis of scE$^2$TM topic and gene embeddings on human pancreas scRNA-seq data. (a) Topic and gene embedding. UMAP visualization shows the global distribution of genes ({\color{myblue2}${\bullet}$}) and topics  ({\color{myred2}${\star}$}) in the learned embedding space. 
    (b) Visualization of the embedding for topic 12 and 69 and their top-$10$ genes.
   (c) Gene Set Enrichment Analysis (GSEA) of topic 12 and 69. Leading-edge analysis was performed on the ``\texttt{REACTOME\_INSULIN\_PROCESSING}'' pathways using Topic 12. The running sum enrichment score is calculated by GSEA. (d) GSEA analysis of topic 69 on ``\texttt{VANGURP\_PANCREATIC\_BETA\_CELL}'' pathway. (e) Topic-gene similarity. Comparison of cosine similarity between each topic and its top genes.}
\label{figure6}  
\end{figure*}

\subsection{scE$^2$TM identifies distinct topics linked to pancreatic function}

We evaluated scE$^2$TM's biological discovery capabilities through analysis of a human pancreas scRNA-seq dataset \cite{wang2016single}. UMAP visualization of gene and topic embeddings shows that scE$^2$TM learns geometrically coherent representations where topics form distinct hubs, surrounded by functionally related genes (Fig. \ref{figure6}a). This clustering pattern, induced by ECR approach, contrasts with methods lacking regularization, which produce diffuse, overlapping gene-topic associations. 
By visualizing topics 12 and 69 and their associated top-$10$ genes, we observe a strong co-localization between genes comprising the ``\texttt{REACTOME\_INSULIN\_PROCESSING}'' and ``\texttt{VANGURP\_PANCREATIC\_BETA\_CELL}'' pathways and their corresponding enriched topics (Fig. \ref{figure6}b).

We investigated the top ten genes for each scE$^2$TM topic and calculated their frequency of occurrence across topics. Our results show the ten most frequent genes, including INS, CEACAM7, and UCA1 (Supplementary Table \ref{supt3}), all experimentally validated or reported in the literature as significantly associated with pancreatic function or related disease mechanisms
\cite{andrali2008glucose,raj2021ceacam7,chen2016long}.
Pathway enrichment analysis identifies many enriched pathways closely related to pancreatic function for each topic (Benjamin-Hochberg (BH) false discovery rate $<$ 0.01; Supplementary Fig. \ref{supf7}). 
Enrichment of ``\texttt{REACTOME\_INSULIN\_PROCESSING}'' and ``\texttt{VANGURP\_PANCREATIC\_BETA\_CELL}'' highlights the model’s ability to capture key pancreatic functions, including insulin biosynthesis and cell-specific molecular characteristics (Fig. \ref{figure6}c,d). 
Cosine similarity analysis quantifies the geometric structure of the embedding space: genes with higher topic weights display greater embedding 
similarity to their corresponding topics (Fig. \ref{figure6}e).
The analysis of human pancreas data establishes that scE$^2$TM recovers biologically coherent topic-gene structures in the embedding space and successfully identifies key pancreatic functional genes and pathways, validating its capability to uncover mechanistically relevant biology from single-cell transcriptomes.

\subsection{Topic-based perturbation reveals distinct interferon response in cell-state transitions}

\begin{figure}
    \centering
    \includegraphics[width=0.9\linewidth]{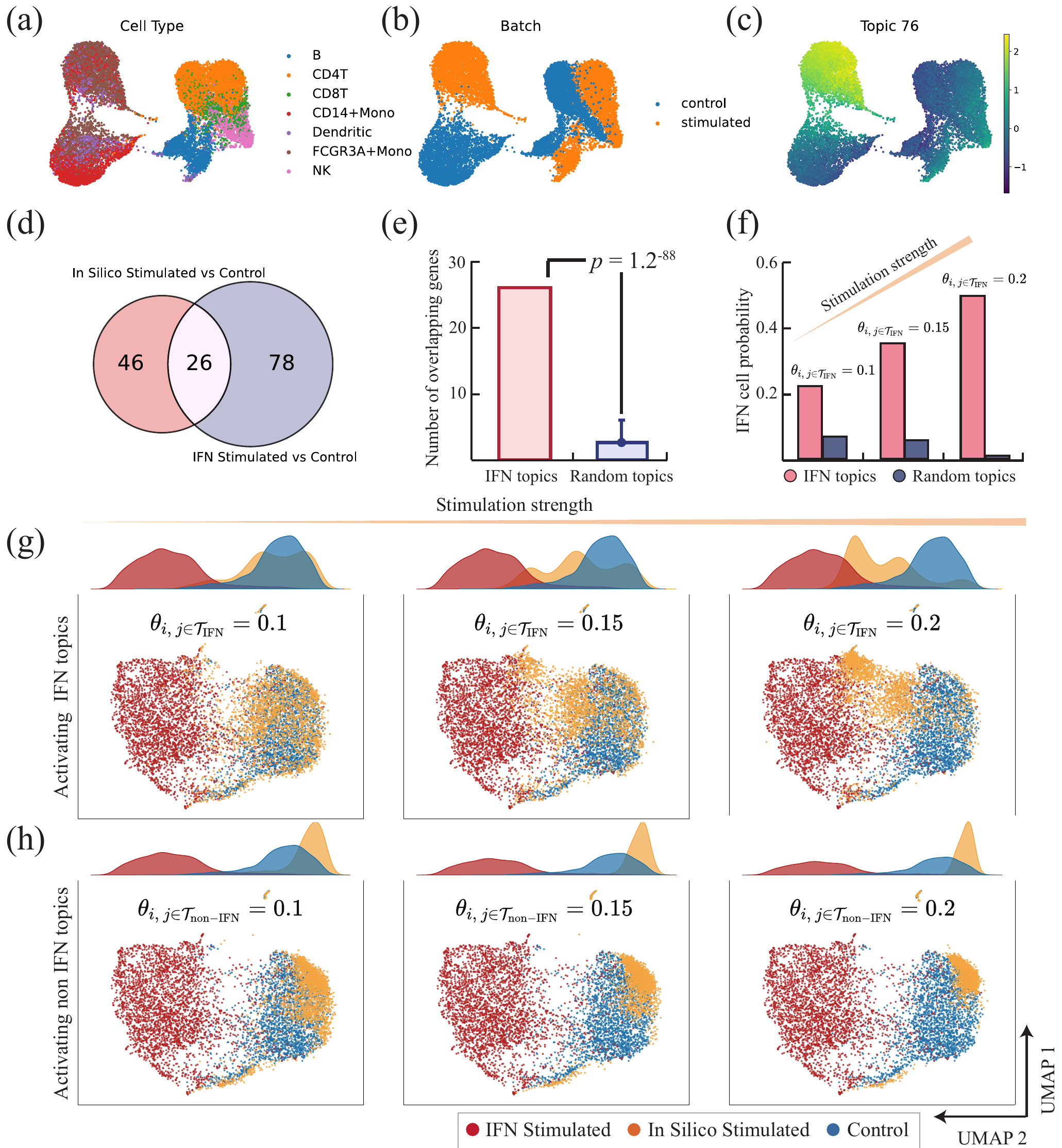}
    \caption{\footnotesize scE$^2$TM predicts cellular response to IFN stimulation. 
    (a) UMAP visualization of PBMC cells colored by cell types. 
    (b) UMAP visualization of PBMC cells colored by treatment conditions.
    (c) UMAP visualization of PBMC cells colored by the scores of one of the IFN-related topics. 
    (d) Consistency of the in-silico IFN stimulation in reference to the experimental IFN response. The Venn diagram shows the overlap of up-regulated genes derived from the in silico stimulation by setting the IFN-specific topic intensity to 0.2 in the unperturbed cells with those up-regulated genes by the experimental IFN-$\beta$ stimulation.
    (e) Statistical significance of the gene overlap from perturbing the IFN topics versus perturbing random topics. 
    We randomly sampled sets of topics matching the number of IFN-specific topics, perturbed these sampled topics, and obtained their overlap with the observed DE genes.
    This was repeated 1,000 times to fit an normal distribution, which was used as the null distribution to compute the $p$-value for the observed overlap.
    (f) Probability of the in-silico IFN stimulated control cells. The dataset was split into training and test sets, and the classification probabilities are generated by a Support Vector Machine (SVM) classifier with Platt scaling on the test set. $\theta_{i,j}$ denotes the topic intensity of cell~$i$ on topic~$j$, and $\mathcal{T}_{\mathrm{IFN}}$ denotes the set of IFN-specific topics.
    (g) Cell representations as a function of perturbation strength for the IFN-specific topics. UMAP for the reconstructed CD4T cells in the control (green), IFN-stimulated (blue), and in silico stimulated (yellow) are shown under weak and strong perturbation strength. The corresponding density plots are displayed to illustrate the proportions of the three cell conditions.
    (h) Cell representations as a function of perturbation strength for the non-IFN-specific topics.
 }
    \label{figure7}
\end{figure}

We applied scE$^2$TM to the PBMC dataset from Kang \textit{et al} \cite{kang2018multiplexed}, which includes single-cell RNA-seq data from lupus patient PBMCs treated with IFN-$\beta$, a type I IFN. 
scE$^2$TM effectively separates known cell types and treatment conditions while
identifying several topics specifically associated with IFN stimulation (Fig. \ref{figure7}a,b). Differential expressed topic analysis (Section \ref{sec:topicdifferential}) identifies topics 59 and 76 (IFN-specific topic) as significantly up-regulated in IFN-stimulated cells versus control cells (BH-adjusted $q$-value $<$ 0.01; Fig. \ref{figure7}c; Supplementary Fig. \ref{supf8}). Topic 76 exhibits the highest activation levels in FCGR3A+Mono cells. The top genes of topic 76 are significantly enriched for for both FCGR3A+Mono-specific up-regulate genes (Supplementary Fig. \ref{supf9}a) and type I IFN response pathway components (Supplementary Fig. \ref{supf9}b). 


We performed a topic-guided perturbation analysis (Section \ref{sec:perturbation}) to investigate whether perturbing relevant topics (i.e., gene expression programs) could induce intended cellular state changes. We identified 104 Differential Expressed Genes (DEGs) significantly up-regulated in IFN-stimulated versus control cells, which serve as a reference gene set. Using the IFN-specific topics 76 and 59,  we extracted cell-topic distributions from control cells and performed \textit{in silico} stimulation by increasing the intensities of these two topics. 
The decoder computes the expected gene expression of both perturbed and unperturbed control cells, revealing 72 DEGs between these conditions, 26 of which overlap with the reference DEGs (Fig. \ref{figure7}d and Supplementary Fig. \ref{supf15}). 
Among these, the most strongly upregulated overlapping gene, CH25H, has been identified as an IFN-dependent gene with potent antiviral activity \cite{liu2013interferon}. Permutation testing shows that this overlap is significantly exceeds expected levels from perturbing random topics (Fig. \ref{figure7}e), confirming the specificity of IFN-related topic perturbations.

Perturbation strengths modulates the magnitude of cell-state transitions. As the intensity of IFN-specific topic perturbation increases, a greater proportion of control cells become identified as IFN-stimulated (Fig. \ref{figure7}f). To visualize these transitions at the population level, we focus on CD4T cells, which represent the major cell type in this dataset. We projected the reconstructed representations of control and IFN-stimulated cells into a UMAP space, and then mapped the perturbed control cells into the same space. With increasing IFN-specific topic stimulation, the perturbed control cells progressively converge toward the actual IFN-stimulated cells (Fig. \ref{figure7}g). 
In contrast, perturbation with a set of non-IFN-specific topics that showed low correlation with IFN and were matched in number to the IFN-specific topics resulted in cells that remained clustered with unperturbed control cells (Fig. \ref{figure7}h). Moreover, perturbing topics 76 and 59 individually did not lead the perturbed control cells to shift noticeably closer to the actual IFN-stimulated cells (Supplementary Fig. \ref{supf10}). 
These perturbation experiments establish that scE$^2$TM-identified topics capture biologically relevant gene programs that can be leveraged to computationally simulate treatment-induced cell-state transitions.

\begin{figure}[b!]
    \centering
    \includegraphics[width=0.95\linewidth]{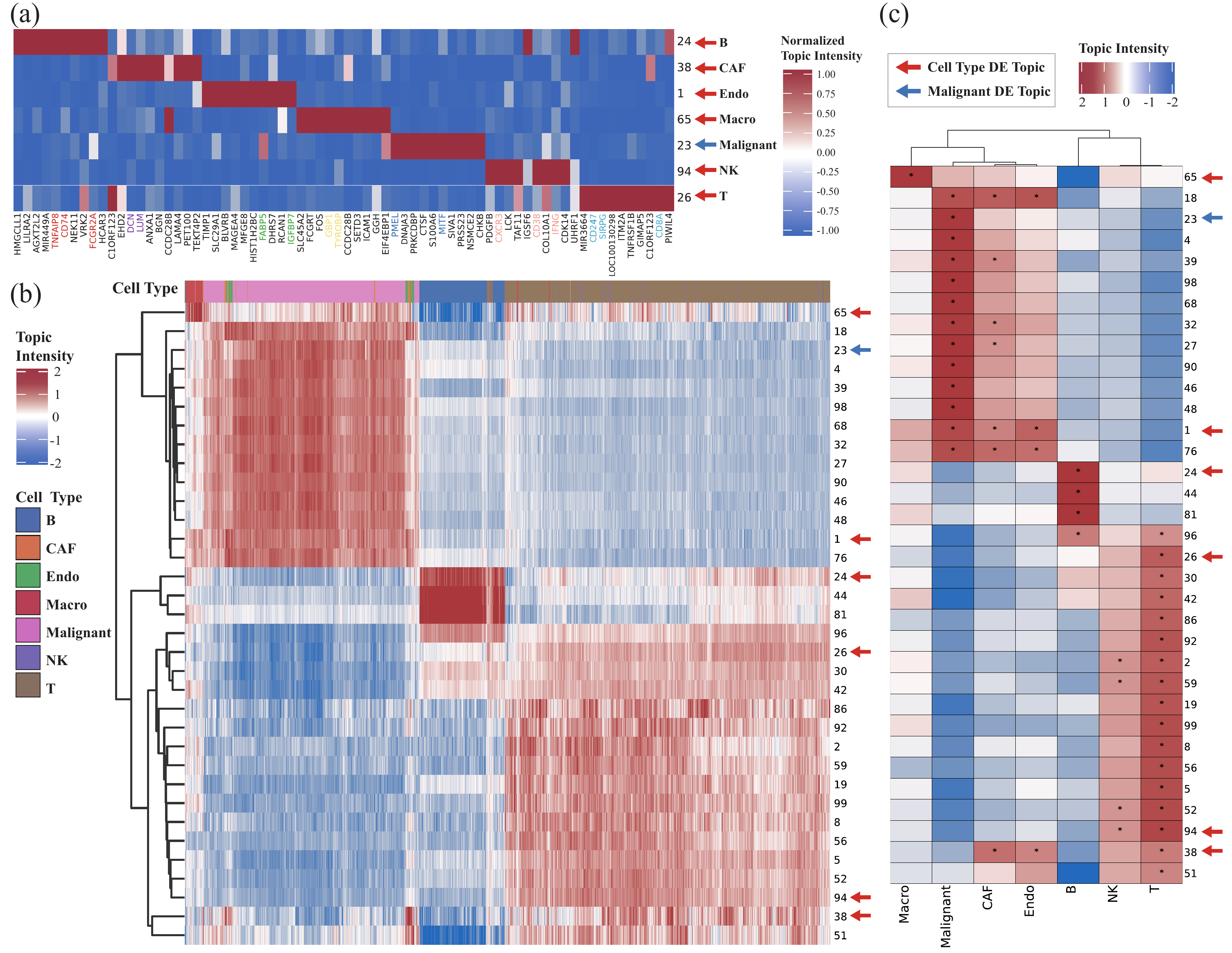}
    \caption{
    Topic analysis of the singe-cell melanoma dataset. 
    (a) Gene-topic heatmap. For visualization purposes,   the scE$^2$TM inferred topic scores were standardized (i.e., mean centered and scaled). Highlighted genes indicate overlap with known marker genes from the CellMarker database, with colors distinguishing cell types. 
    (b) Hierarchical clustering heatmap of cells. The topic strengths displayed here represent the latent Gaussian mean without softmax application. The most highly variable topics were shown. For the representative topics indicated by arrows, their top gene were shown in panel a.
    (c) Differential expression analysis by topics. Colors indicate mean differences between cell group with the target label (cell types or malignant state) and the rest of the cell. Asterisks denote BH-adjusted $q$-values $<$ 0.01.} 
    \label{figure81}
\end{figure}

\subsection{Topic-based perturbation shifts melanoma cells towards normal state}
\begin{figure}[t!]
    \centering
    \includegraphics[width=\linewidth]{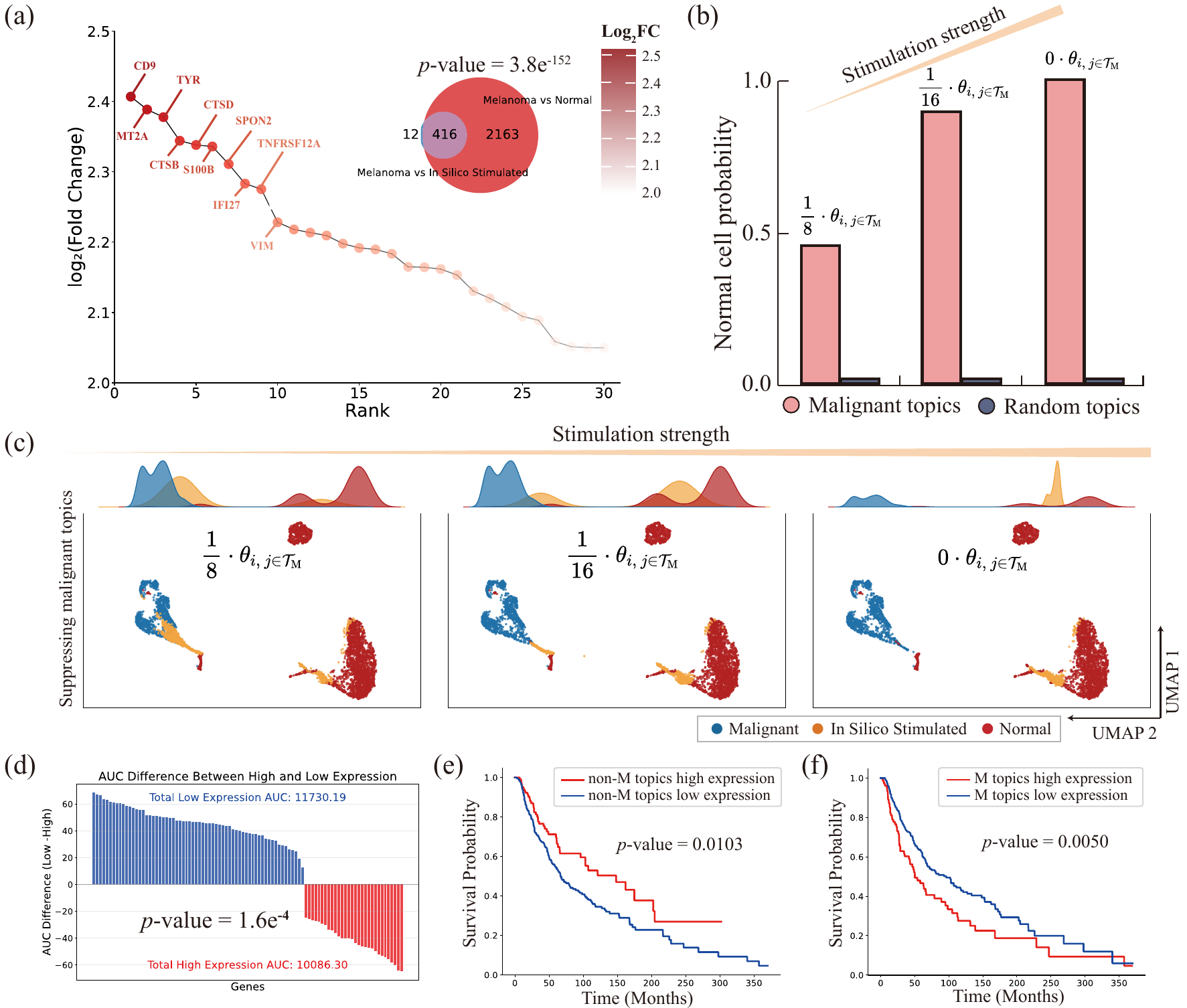}
    \caption{\small Topic-driven perturbation responses and prognostic associations in melanoma.
    (a) Consistency between simulated topic stimulation and malignant cell response.
    The Venn diagram illustrates the overlap between genes up-regulated in malignant cells relative to in silico stimulation (malignant-specific topic intensity in malignant cells set to 0) and genes up-regulated relative to normal cells. The $p$-value reflects the probability of obtaining an overlap as large as the observed value under the null distribution generated by random topic perturbations. The hockey stick plot displays the overlapping genes sorted by degree of up-regulation. 
    (b) Normal cell classification probabilities as a function of perturbation strength. The probability of malignant cells being classified as normal after perturbing malignant-specific topics versus perturbing random topics. The probability was computed by a separate SVM classifier trained to predict discriminate normal from malignant. 
    (c) Perturbed cell-state visualization. UMAP projections and density maps of reconstructed cells under different stimulus intensities. $\mathcal{T}_{\mathrm{M}}$ denotes the set of malignant-specific topics.
    (d) Gene survival analysis. Gene-level survival analysis of the 416 overlapping genes shown in (a) using an unseen TCGA melanoma cohort. Each bar represents the AUC difference between patient groups stratified by high versus low gene expression levels.
    (e) Survival analysis for non-malignant (non-M) topics in the TCGA melanoma cohort.
    (f) Survival analysis for malignant-specific (M) topics in the TCGA melanoma cohort.
    }
    \label{figure9}
\end{figure}

We applied scE$^2$TM to a melanoma scRNA-seq dataset \cite{tirosh2016dissecting,chen2024comprehensive} to identify gene topics associated with malignancy. The dataset comprised 4,645 malignant, immune, and stromal cells isolated from 19 melanoma patients, including ten metastases to lymphoid tissues, eight distant-site metastases, and one primary acral melanoma. Differential topic analysis reveals 33 malignant-specific topics in comparison with normal cell states (Supplementary Fig. \ref{supf11}). These topics exhibit high selectivity for cell-type marker genes and  strong discriminative power for distinguishing cell types (Fig. \ref{figure81}). We performed topic-based perturbation by reducing the topic scores for malignant-specific topics and computed the expected gene expression for perturbed and unperturbed malignant cells using the decoder. DEG analysis identifies 428 genes that are significantly upregulated in reconstructed unperturbed malignant cells compared with perturbed malignant cells (Supplementary Fig. \ref{supf12}). As reference, we also identified DEGs using the observed malignant and normal cells. The two differential expressed gene sets show significant overlaps (Fig. \ref{figure9}a). 

To assess whether the perturbed malignant cells shift towards the normal state, we trained an SVM classifier to predict cell states (malignant or normal) (Section \ref{sec:svm}).
As perturbation intensity increases by progressive suppression of malignant-specific topics, a greater proportion of malignant cells become classified as normal (Fig. \ref{figure9}b), indicating that these topics are critical for maintaining the malignant phenotype and might represent key drivers of malignant transformation. 
To visualize these transitions, we projected perturbed malignant cells into UMAP space. With increasing perturbation of the malignant-specific topics, the perturbed cells gradually converge toward the normal cell cluster (Fig. \ref{figure9}c), whereas perturbing non-malignant-specific topics results in cells remaining clustered among the original malignant cells (Supplementary Fig. \ref{supf13}). 
Furthermore, we found that perturbing the top-$10$ malignant-specific topics induced detectable shifts in cellular states, and that these shifts became more pronounced as the number of perturbed malignant-specific topics increased (Supplementary Fig. \ref{supf14}).
This observation indicates that the malignant-specific topics identified by scE$^2$TM represent key transcriptional axes of tumor progression, and perturbing them reverses malignant cells toward normal phenotypes.

To assess the clinical relevance of our analysis, we further analyzed melanoma patient data from The Cancer Genome Atlas (TCGA) \cite{hoadley2018cell}. The up-regulated genes identified in the melanoma dataset show a significant negative correlation with survival rates in TCGA melanoma cohort patients (Fig. \ref{figure9}d; $p$-value = 1.6e-4), confirming that perturbation of the target indeed impacts key signaling pathways in melanoma biology.
We then applied the melanoma-trained scE$^2$TM model to the TCGA melanoma gene expression data and examined the associations between the previously identified malignant-specific topics and patient outcomes. The expression of malignant-specific topics shows significant associations with survival (Fig. \ref{figure9}e,f), demonstrating that scE$^2$TM-identified topics capture clinically relevant gene programs that generalize across independent patient cohorts. 
Together, scE$^2$TM reveals malignant-specific topics and can be robustly extended to independent TCGA melanoma studies. These findings suggest that interpretable topic characterization holds promise as a bridge linking single-cell transcriptomics to cancer clinical prognosis prediction.

\section{Discussion}
Identifying marker genes for previously uncharacteristic cell types and disease-specific transcriptional states remains a key challenge in scRNA-seq analysis. Topic models present an effective framework that can identify interpretable gene-topic distributions and characterize cell by these topics. However, existing topic models suffer from the lack of external knowledge and interpretation collapse, where many topics convey redundant information. To this end, we present scE$^2$TM, a novel single-cell embedding topic model through two technical contributions. 

First, we integrate external knowledge from a pretrained foundation model scGPT through a novel cross-view distillation strategy. This approach effectively coordinates cellular neighborhood information across different representations, encourages consistent topic assignment between cells and their corresponding neighbors in the embedding space, thereby integrating rich contextual information while maintaining computational efficiency. Our results demonstrate that this strategy achieves both superior clustering performance and enhanced scalability, enabling direct knowledge transfer from pretrained foundation models without requiring joint clustering or label transfer operations. 

Second, to address interpretation collapse, we adopt ECR, which treats topics as clustering centroids and genes as samples. Through optimal transport, the ECR ensures that each topic serves as the anchors of a distinct gene cluster. Quantitative experiments show scE$^2$TM's substantial improvement over benchmark models, with increases of 8.6\% in interpretation purity, 66.5\% in topic quality, 27.5\% in GSEA quality, and 16.1\% in ORA quality. Qualitative experiments further confirm that scE$^2$TM topics capture cell type-specific and functionally relevant biological signals. Our correlation analysis between clustering and interpretability metrics reveals that high clustering performance does not guarantee biologically meaningful interpretations, underscoring the importance of dedicated interpretability assessment.

Beyond benchmarking interpretability, our topic-based perturbation experiments in PBMC and melanoma datasets demonstrate scE$^2$TM’s potential to model cell-state transitions \textit{in silico}. By perturbing topic activations associated with IFN response or malignant programs, scE$^2$TM accurately reconstructed trajectories from stimulated to resting immune states and from tumor to normal-like melanoma cells, respectively. These findings highlight that topic perturbations capture causal relationships between gene programs and cellular phenotypes. In practical terms, this framework can support \textit{in silico} hypothesis testing—predicting how cells might respond to immune stimulation, drug treatment, or oncogenic suppression—thus offering a powerful tool for precision medicine and therapeutic target discovery in real-world clinical settings.

Future extensions of our study will focus on three directions. First, we will collect tissue- or cell type-specific pathway datasets to enable a more accurate evaluation of the model’s interpretability. Although scE$^2$TM identifies topics enriched for diverse and coherent biological pathways, several interpretability metrics (e.g., ORA$_Q$ and GSEA$_Q$) are evaluated against all known pathways and therefore cannot determine how well these pathways reflect the underlying cellular characteristics. Collecting biological pathway datasets with tissue- or cell type-specific information is a promising direction for improving our interpretability evaluation benchmarks.
Nevertheless, we anticipate that diversity and consistency will remain valid assessment criteria.
Secondly, existing single-cell embedding topic models generate global topic-gene dependencies for entire datasets, potentially overlooking cell-specific characteristics. Developing hierarchical topic structures that generate topics at multiple granularities would enable more refined interpretations focused on individual cell features \cite{wu2024survey}.
Finally, with the increasing prevalence of multimodal single-cell assays, extending scE$^2$TM and our interpretability metrics to multimodal data would enable more comprehensive biological signal capture across modalities \cite{ZHOU2023100563,coleman2025resolving,cui2025towards}.

In summary, scE$^2$TM represents an external knowledge-guided single-cell embedded topic model with robust clustering performance and enhanced interpretability. We identified critical limitations in the current evaluation framework and introduced a suite of quantitative metrics to address them. We envision that these contributions will advance our understanding of cellular biology and establish foundations for future developments in interpretable single-cell analysis.

\section{Methods}


\subsection{scRNA-seq datasets}
Twenty scRNA-seq datasets were used to conduct experiments, including human and mouse data.
For all single-cell benchmarking datasets, we performed  the same preprocessing steps using Scanpy package (v.1.9.8). Specifically, we used ``\texttt{scanpy.pp.log1p()}'' to log-transform the counts of each cell, and then use ``\texttt{scanpy.pp.highly\_variable\_genes()}'' to select the top-$5000$ highly variable genes.    
Details of all single-cell data are provided in Supplementary Table \ref{supt4}.   

In addition, we used two scRNA-seq datasets for the topic-based perturbation experiments. The first dataset by Kang \textit{et al} \cite{kang2018multiplexed} consists of two groups of PBMCs, one control and one stimulated with IFN-$\beta$. We applied the same preprocessing steps as in previous work \cite{seninge2021vega}, ultimately yielding a dataset containing 18,868 cells. Raw data are available at GSE96583. 
The second dataset were obtained from Tirosh \textit{et al} \cite{tirosh2016dissecting} (GSE72056) for melanoma study. We utilized the dataset processed by Chen \textit{et al}\cite{chen2024comprehensive}, which removed all samples of indeterminate type and resulted in a final dataset of 4,092 cells comprising seven distinct cell types.

\subsection{Clustering benchmark metrics}\label{sec:cluster-metric}
For evaluating cell clustering, we used standard metrics, namely the Normalized Mutual Information (NMI) and the Adjusted Rand Index (ARI). 
All embedding maps were visualized using UMAP. We clustered the cells using Louvain’s algorithm. Furthermore, to better assess the clustering performance of cell embeddings, we experimented with multiple resolution values and reported the results with the highest ARI of each method, following previous studies \cite{luecken2022benchmarking,chen2024comprehensive}. Moreover, we assessed model interpretability using the proposed metrics: IP, TC, TD, TQ, ORA$_N$, ORA$_U$, ORA$_Q$, GSEA$_N$, GSEA$_U$, and GSEA$_Q$.

\subsection{Baselines methods}
For the single-cell embedded topic model, we used scVI-LD \cite{svensson2020interpretable}, scETM \cite{zhao2021learning}, 
SPRUCE \cite{subedi2023single}, and d-scIGM \cite{chen2024comprehensive} as baselines. The scVI-LD method provided interpretability by determining the relationship between cell representative coordinates and gene weights through matrix factorization, where the cell representative coordinates are considered as topics. Notably, d-scIGM utilized hierarchical prior knowledge to bootstrap the model, which may lead to data leakage in the interpretability evaluation. To prevent that, we used the version without prior knowledge described in the original article. 

All experiments on the benchmark models were performed using publicly available code. We used the optimal hyperparameters reported in their original papers.
For the models that utilize information about the topology of the cellular network, we used scTAG \cite{yu2022zinb} and scLEGA \cite{liu2024sclega} as baselines. In addition, scVI \cite{lopez2018deep} was also used as a baseline model, which is a classical single-cell deep embedding method that has demonstrated excellent overall performance in previous comprehensive evaluations \cite{luecken2022benchmarking}.
More detailed descriptions of the relevant methods can be found in Supplementary Note 1.

\subsection{Quantitative assessment of interpretability}\label{sec:interpre-metric}
\subsubsection{Topic coherence}
Topic Coherence (TC) quantifies the semantic consistency of a topic by measuring the statistical co-occurrence patterns of its characteristic terms. For a given topic, TC evaluates whether its top-associated terms frequently co-appear in relevant contexts within an external corpus, indicating a coherent semantic representation. This metric strongly correlates with human interpretability judgments in traditional topic modeling \cite{lau2014machine}. We adapt TC to single-cell biology by leveraging biological pathways as semantic units, using the GASE database \cite{gillespie2022reactome} as the external knowledge base. Specifically, we suggest that genes assigned to the same topic collaboratively participate in common biological pathways. Formally, TC is defined as
\begin{equation}
\mathrm{TC}=\frac{1}{K} \sum_{k=1}^{K} \frac{2}{h(h-1)} \sum_{i=1}^{h} \sum_{j=i+1}^{h} F\left(\mathbf{g}_{i}^{(k)}, \mathbf{g}_{j}^{(k)}\right),
\end{equation}
where $\left\{\mathbf{g}_{1}^{(k)}, \ldots, \mathbf{g}_{h}^{(k)}\right\}$ denotes the top-$h$ most likely genes in the $k$-th topic. And the correlation of word pairs is estimated as follows:
\begin{equation}
F\left(\mathbf{g}_{i}, \mathbf{g}_{j}\right)=\frac{\log \frac{P\left(\mathbf{g}_{i}, \mathbf{g}_{j}\right)}{P\left(\mathbf{g}_{i}\right) P\left(\mathbf{g}_{j}\right)}}{-\log \left(P\left(\mathbf{g}_{i}, \mathbf{g}_{j}\right)\right)},
\end{equation}
where $P\left(\mathbf{g}_{i},\mathbf{g}_{j}\right)$ is the probability that genes $\mathbf{g}_{i}$ and $\mathbf{g}_{j}$ co-occur in the same biological process, and $P\left(\mathbf{g}_{i}\right)$  is the marginal probability of gene $\mathbf{g}_{i}$. TC ranges continuously on [0, 1], with higher values indicating superior coherence.

\subsubsection{Topic diversity}
Topic Diversity (TD) quantifies the distinctness of biological mechanisms captured by different topics. Existing single-cell embedded topic models often suffer from interpretation collapse, i.e., the model mines many overlapping topics. Therefore, topic diversity should be one of the important metrics for assessing interpretability. Formally, TD is defined as follows:
\begin{equation}
\mathrm{TD}=\frac{|\mathrm{Unique}(M)|}{|M|},
\end{equation}
where $M=\bigcup_{k=1}^K\left\{\mathbf{g}_{1}^{(k)}, \ldots, \mathbf{g}_{h}^{(k)}\right\}$ denotes the gene pool containing the top-$h$ genes corresponding to all topics. $|M|$ and $|\mathrm{Unique}(M)|$ denote the number of all genes and the number of unique genes, respectively. TD ranges continuously over [0, 1], where $\mathrm{TD} \to 0$ indicates high redundancy and $\mathrm{TD} \to 1$ reflects maximal diversity.

\subsubsection{Topic quality}
There is a mutual trade-off between topic coherence and topic diversity in the text mining domain \cite{lin-etal-2024-hierarchical}, and we suggest that this phenomenon also exists in the single-cell domain. Since some important genes correspond to multiple gene programs, these genes may appear in different topics at the same time, which would result in a higher TC score but a slightly reduced TD score. In addition, specific gene programs may be captured repeatedly by topics, which would further exacerbate the above problem. Therefore, we define the product of topic diversity and topic coherence as the overall Topic Quality (TQ) following Dieng \textit{et al} \cite{lin-etal-2024-hierarchical}, i.e.,
\begin{equation}
\mathrm{TQ}=\mathrm{TC} \times \mathrm{TD},
\end{equation}
which ranges from 0 to 1, with higher values indicating superior balance between coherent biological pathways and diverse functional coverage.

\subsubsection{Interpretation purity}
Interpretation Purity (IP) captures the intuition that cells assigned to the same topic should share common biological characteristics. Specifically, it tests the hypothesis that dominant topic assignments should align with known cell type annotations at a coarse-grained level, reflecting shared functional states. Adapted from the cluster purity metric, IP quantifies this alignment as:
\begin{equation}
\mathrm{IP}(\Omega, \mathbf{Y}):=\frac{1}{n} \sum_{k=1}^{K} \max _{j}\left|\Omega_{k} \cap \mathbf{Y}_{j}\right|,
\end{equation}
where $\Omega = \{\Omega_{k} \mid k=1, \ldots, K\}$ denotes the topic assignments with $\Omega_{k} = \{ i \mid \arg\max_j \theta_{ij} = k,\; i=1, \ldots, n \}$ representing cells assigned to topic $k$. Besides, $\mathbf{Y} = \{\mathbf{Y}_j \mid j=1, \ldots, J\}$ represents $J$ ground truth cell types with $\mathbf{Y}_j$ containing cells of type $j$. And $n$ is the total number of cells. The metric ranges within $[0, 1]$, where higher values indicate stronger agreement between topics and biological cell states, reflecting more meaningful cellular interpretations.

\subsubsection{Quality of topic over-representation analysis}
Complementing the cell-level assessment of interpretation purity, which measures the alignment of topics with biological cell states, we now evaluate the functional relevance of topics at the gene and pathway level. To validate the biological mechanisms captured by the mined topics, we perform Over-Representation Analysis (ORA) that statistically evaluates whether top topic-associated genes cluster significantly in known biological pathways. This quantifies how effectively topics capture interpretable biological mechanisms. Specifically, we propose three metrics that assess different dimensions of this functional enrichment quality: (i) number of pathways enriched by topic ORA (ORA$_N$), which measures the breadth of distinct biological functions captured across all topics; (ii) uniqueness of pathways enriched by topic ORA (ORA$_U$), which quantifies the avoidance of pathway repetition across topics; and (iii) quality of topic ORA (ORA$_Q$) balances the assessment in both coverage and uniqueness. The formal definitions of these metrics are as follows:
\begin{equation}
\mathrm{ORA}_N=|\mathrm{Unique}(P_O)|,\mathrm{ORA}_U=\frac{|\mathrm{Unique}(P_O)|}{|P_O|},\mathrm{ORA}_Q=\mathrm{ORA}_N \times \mathrm{ORA}_U,
\end{equation}
where $P_{O}$ contains the topic enrichment pathways obtained via ORA. $|P_{O}|$ and $|\mathrm{Unique}(P_{O})|$ denote the number of all pathways and the number of unique pathways, respectively. Similar to TQ, $\mathrm{ORA}_Q$ is defined as the product of ORA$_N$ and ORA$_U$.

\subsubsection{Quality of topic gene set enrichment analysis}
While ORA evaluates pathway enrichment using thresholded top genes, this approach requires arbitrary selection of gene cutoffs \cite{zhao2021learning,chen2024comprehensive}. To overcome this limitation, we conduct GSEA, a threshold-free method that evaluates enrichment across the entire ranked list of genes sorted by their association strength with each topic. GSEA computes a running enrichment score that avoids discrete gene selection, providing more robust pathway associations (method details in Supplementary Note 2). Following our ORA evaluation framework, we adapt three analogous metrics (GSEA$_N$, GSEA$_U$, GSEA$_Q$) to quantify biological interpretability:
\begin{equation}
\mathrm{GSEA}_N=|\mathrm{Unique}(P_{G})|,\mathrm{GSEA}_U=\frac{|\mathrm{Unique}(P_{G})|}{|P_{G}|},\mathrm{GSEA}_Q=\mathrm{GSEA}_N \times \mathrm{GSEA}_U,
\end{equation}
where $P_{G}$ contains the topic enrichment pathways obtained via GSEA.

\subsection{scE$^2$M method details}

\subsubsection{Integrating external embedding via cross-view knowledge distillation}
\label{sec:exk}
scE$^2$TM is a Variational AutoEncode (VAE) framework. The topic-based encoder maps the input sample into the latent Gaussian distribution. The reparameterized sample is then transformed via softmax function to represent the cell-topic latent variable $\boldsymbol{\theta}$. 
To integrate internal embedding learned directly from the scRNA-seq training data with the external knowledge provided by the single-cell foundation model, we propose a cross-view distillation strategy.
The key idea is that the same cell from different perspectives should be mapped to the same topic, and that consistent topic assignment between each cell and the corresponding neighboring cell in another perspective is encouraged by integrating the cell's neighborhood information. 
Specifically, let $\mathbf{x}$ and $\mathbf{v}$ be the internal and external cell embedding representations, respectively. We introduce a cell-topic head 
$f(\cdot): \mathbf{x} \rightarrow \boldsymbol{\theta} \in \mathbb{R}^{K}$ and a cell-cluster head $g(\cdot): \mathbf{v} \rightarrow \boldsymbol{\varphi} \in \mathbb{R}^{K}$, where $K$ is the number of target topics or clusters\footnote{Clusters and topics are used to mainly distinguish the internal and external embedding, respectively.}. For $i\in \{1,\ldots,n\}$ cells, we denote the soft topic/cluster assignment of cell $i$ as
\begin{align}
    \boldsymbol{\mu}, \log\boldsymbol{\sigma} &= f(\mathbf{x}_{i}) = \mathrm{MLP}\left(\mathbf{x}_{i}; \mathbf{W}_f\right)\\
    \mathbf{z}_{i} &\sim \mathbf{\boldsymbol{\mu}} + \boldsymbol{\sigma}\mathcal{N}(0, \mathbf{I})\\
    \boldsymbol{\theta}_{i} &= \mathrm{Softmax}\left(\mathbf{z}_i\right),\\
    \boldsymbol{\varphi}_{i} &= g\left(\mathbf{v}_{i}\right) = \mathrm{Softmax}\left(\mathrm{MLP}\left(\mathbf{v}_{i}; \mathbf{W}_g\right))\right.
\end{align}
where the softmax function $\exp(z_k)/\sum_k\exp(z_k)$ is applied after the MLP function to produce the probabilistic topic and cluster assignments from the two views, respectively.

To encourage topic consistency between internal and external perspectives of the same cell, we define a cross-view assignment consistency loss:
\begin{equation}
\mathcal{L}_{\mathrm{CON}}=-\log \sum_{i=1}^{n} \boldsymbol{\theta}_{i}^{\top} \boldsymbol{\varphi}_{i},
\end{equation}
which is minimized for similar embedding between $\boldsymbol{\theta}_{i}$ and $\boldsymbol{\varphi}_{i}$. 

To utilize the cell neighborhood information of different views, we first obtain the nearest neighbors set of $\mathbf{x}_{i}$ and $\mathbf{v}_{i}$, denoted as $\mathcal{N}\left(\mathbf{x}_{i}\right)$ and $\mathcal{N}\left(\mathbf{v}_{i}\right)$, respectively. For randomly sampled neighbors $\mathbf{x}_{i}^{\mathcal{N}}\in\mathcal{N}\left(\mathbf{x}_{i}\right)$ and $\mathbf{v}_{i}^{\mathcal{N}}\in\mathcal{N}\left(\mathbf{v}_{i}\right)$, their corresponding topic assignments are defined as 
\begin{equation}
    \boldsymbol{\theta}_{i}^{\mathcal{N}} = f\left(\mathbf{x}_{i}^{\mathcal{N}}\right),\quad
    \boldsymbol{\varphi}_{i}^{\mathcal{N}} = g\left(\mathbf{v}_{i}^{\mathcal{N}}\right). 
\end{equation}

Let $\boldsymbol{\theta}_{ik}$, $\boldsymbol{\theta}_{ik}^{\mathcal{N}}$, $\boldsymbol{\varphi}_{ik}$, $\boldsymbol{\varphi}_{ik}^{\mathcal{N}}$ be the assignment of the $k$-th topic/cluster, the loss terms of the cross-view neighborhood consistency are defined as follows:
\begin{align}
    \mathcal{L}_{\mathrm{NEI}} &= \frac{1}{n} \sum_{i=1}^{n} \sum_{k=1}^{K} (L_{ik}^{\mathbf{x} \rightarrow \mathbf{v}}+L_{ik}^{\mathbf{v} \rightarrow \mathbf{x}}), \\
    L_{ik}^{\mathbf{x} \rightarrow \mathbf{v}} &= -\log \frac{\exp\left(\mathrm{sim}\left(\boldsymbol{\varphi}_{ik}, \boldsymbol{\theta}_{ik}^{\mathcal{N}}\right) / 2\right)}{\sum_{j} \exp{\left(\mathrm{sim}\left(\boldsymbol{\varphi}_{ik}, \boldsymbol{\theta}_{ij}^{\mathcal{N}}\right) / 2\right)}+\sum_{j \neq k} \exp{\left(\mathrm{sim}\left(\boldsymbol{\varphi}_{ik}, \boldsymbol{\varphi}_{ij}\right) / 2\right)}}, \\
    L_{ik}^{\mathbf{v} \rightarrow \mathbf{x}} &= -\log \frac{\exp{\left(\mathrm{sim}\left(\boldsymbol{\theta}_{ik}, \boldsymbol{\varphi}_{ik}^{\mathcal{N}}\right) / 2\right)}}{\sum_{j} \exp{\left(\mathrm{sim}\left(\boldsymbol{\theta}_{ik}, \boldsymbol{\varphi}_{ij}^{\mathcal{N}}\right) / 2\right)}+\sum_{j \neq k} \exp{\left(\mathrm{sim}\left(\boldsymbol{\theta}_{ik}, \boldsymbol{\theta}_{ij}\right) / 2\right)}}
\end{align}

Notably, $\mathcal{L}_{\mathrm{NEI}}$ encourages consistent topic assignment between a cell and its corresponding neighbor in another view, where $L_{ik}^{\mathbf{x} \rightarrow \mathbf{v}}$ aligns the internal feature of the cell to the neighborhood of its external features, and $L_{ik}^{\mathbf{v} \rightarrow \mathbf{x}}$ vice versa. In addition, the second terms of the denominator in the definitions of $L_{ik}^{\mathbf{x} \rightarrow \mathbf{v}}$ and $L_{ik}^{\mathbf{v} \rightarrow \mathbf{x}}$ can minimize inter-topic similarity, resulting in more discriminative topics.  
As a classical neighbor search algorithm, the $k$-nearest neighbor algorithm is susceptible to ``\texttt{dimensionality catastrophe}'' and pivot effect, so we choose mutual kNN to find the set of nearest neighbors of cells \cite{dalmia2021clustering}.
In practice, we set the number of nearest neighbors $|\mathcal{N}\left(\mathbf{x}_{i}\right)|=|\mathcal{N}\left(\mathbf{v}_{i}\right)|$ = 15 for $i \in \{1, \ldots, n\}$ for all datasets.

To prevent all samples from collapsing into a few topics, we introduce an entropy regularization term to stabilize the training process, which is defined as follows:
\begin{equation}
    \mathcal{L}_{\mathrm{REG}}=-\sum_{i=1}^{n}\sum_{k=1}^{K}\left(\boldsymbol{\theta}_{ik} \log \boldsymbol{\theta}_{ik}+\boldsymbol{\varphi}_{ik} \log \boldsymbol{\varphi}_{ik}\right).
\end{equation}

Finally, we derive the overall loss function of our cross-view distillation strategy as follows:
\begin{equation}
    \mathcal{L}_{\mathrm{CVE}}=\mathcal{L}_{\mathrm{CON}}+\mathcal{L}_{\mathrm{NEI}}-\alpha \cdot \mathcal{L}_{\mathrm{REG}}.
    \label{eq:l-cme}
\end{equation}

\subsubsection{Embedding clustering regularization (ECR)}
\label{sec:ecr}
To mitigate the potential interpretation collapse problem, we extend the ECR approach of Wu \textit{et al} \cite{wu2023effective} to the single-cell domain to model the relationship between topics and genes.
In our context, each topic serves as a centroid of a distinct gene cluster. We adopt ECR to model the soft assignment of genes to topics using a transport scheme defined specifically for the optimal transport problem. Specifically, $K$ topic embeddings serve as cluster centroids and $V$ gene embeddings represent the data points. Let $n_k$ denote the number of genes assigned to topic embedding $\mathbf{t}_k$, and $s_k = n_k / V$ represent the corresponding cluster size proportion. The vector $\mathbf{s} = (s_1, \ldots, s_K)^{\top} \in \Delta_K$ summarizes all cluster size proportions. We initially assume uniform distribution over teh cluster size $n_{k}=V / K$, $\mathrm{s}=(1 / K, \ldots, 1 / K)^{\mathrm{T}}$, which is shown to  avoid trivial solutions with empty clusters \cite{wallach2009rethinking,wu2023effective}. 
For the learning objective, we define two discrete measures of gene embedding $\mathbf{g}_{m}$ and topic embedding
$\mathbf{
t}_{k}$: $\gamma=\sum_{m=1}^{V} \frac{1}{V} \delta_{\mathbf{g}_{m}}$ and $ \phi=\sum_{k=1}^{K} s_{k} \delta_{\mathbf{t}_{k}}$, where $\delta_z$
denotes the Dirac unit mass on $z$. ECR formulates the entropic regularized optimal transport between $\gamma$ and $\phi$ as
\begin{align}
    \underset{\boldsymbol{\pi} \in \mathbb{R}_{+}^{V \times K}}{\arg \min } \mathcal{L}_{\mathrm{OT}_{\varepsilon}}(\gamma, \phi) &= \sum_{m = 1}^{V} \sum_{k = 1}^{K}\mathrm{D}\left(\mathbf{g}_{m},\mathbf{t}_{k}\right) \pi_{m k}+\sum_{m = 1}^{V} \sum_{k = 1}^{K} \varepsilon \pi_{m k}\left(\log \left(\pi_{m k}\right)-1\right)\notag\\ 
    &\text { s.t. } \quad \boldsymbol{\pi} \mathbf{1}_{K} = \frac{1}{V} \mathbf{1}_{V} \quad \text { and } \quad \boldsymbol{\pi}^{\top} \mathbf{1}_{V} = \frac{1}{K} \mathbf{1}_{K}
    \label{eq:l-ot}
\end{align}
where the first term is the original optimal transport problem and the second term with the hyperparameter $\varepsilon$ is entropy regularization.
Eq. (\ref{eq:l-ot}) is to find the optimal
transport plan $\boldsymbol{\pi}_{\varepsilon}^{*}$
that minimizes the total cost of transporting weight from gene embeddings to topic embeddings. We measure the transport cost between gene $\mathbf{g}_{m}$ and topic $\mathbf{t}_{k}$ by Euclidean distance: $\mathrm{D}\left(\mathbf{g}_{m},\mathbf{t}_{k}\right)=C_{m k}=\left\|\mathbf{g}_{m}-\mathbf{t}_{k}\right\|^{2}$, and the transport cost matrix is denoted as $ \mathbf{C} \in \mathbb{R}^{V \times K}$. The two constraints in Eq. (\ref{eq:l-ot}) restrict the
weight of each gene embedding $\mathbf{g}_{m}$ as $\frac{1}{V}$
and each topic embedding $\mathbf{t}_{k}$ as $\frac{1}{K}$, where $\mathbf{1}_{K}$ ($\mathbf{1}_{V}$) is a $K(V)$ dimensional column vector of ones. Besides, $\pi_{m k}$ denotes the transport weight from $\mathbf{g}_{m}$ to $\mathbf{t}_{k}$ and $\boldsymbol{\pi} \in \mathbb{R}_{+}^{V \times K}$ is the transport plan that includes the transport weight of each gene embedding to fulfill the weight of each topic embedding.
To meet these constraints, ECR models clustering soft assignments with the optimal transport plan $\boldsymbol{\pi}_{\varepsilon}^{*}$, i.e., the soft-assignment of $\mathbf{g}_{m}$ to $\mathbf{t}_{k}$ is the transport weight between them, $\pi_{\varepsilon, m k}^{*}$. The formula is defined as follows:
\begin{align}
    \mathcal{L}_{\mathrm{ECR}} &= \sum_{m=1}^{V} \sum_{k=1}^{K}\left\|\mathbf{g}_{m}-\mathbf{t}_{k}\right\|^{2} \pi_{\varepsilon, j k}^{*}\label{eq:l-ecr}
\end{align}
where
\begin{align}
    \pi_{\varepsilon}^{*} \leftarrow \operatorname{sinkhorn}(\gamma, \phi, \varepsilon) \approx \underset{\boldsymbol{\pi} \in \mathbb{R}_{+}^{V \times K}}{\arg \min } \mathcal{L}_{\mathrm{OT}_{\varepsilon}}(\gamma, \phi) 
\end{align}
The Sinkhorn algorithm \cite{cuturi2013sinkhorn} is used to compute $\pi_{\varepsilon}^{*}$, which can be efficiently executed on GPUs.



\subsubsection{Sparse linear decoder}
We compute the expected gene expression by taking the dot product of the cell-topic matrix $\boldsymbol{\theta}_{i}$ and gene-topic matrix $\mathbf{B} \in \mathbb{R}^{V \times K}$: $\hat{\mathbf{x}}_i \propto \boldsymbol{\theta}_{i}\mathbf{B}^\top$. 
Prior work has typically modeled $\mathbf{B}$ as the product of topic and gene embeddings \cite{zhao2021learning}. Here we propose to directly derive $b_{mk} \in \mathbf{B}$ from the ECR-guided embedding distance between the gene $m$ and topic $k$: 
\begin{align}\label{eq:gene-topic}
    b_{m k}=\frac{\exp({-\left\|\mathbf{g}_{m}-\mathbf{t}_{k}\right\|^{2} / \tau)}}{\sum_{j=1}^{K} \exp({-\left\|\mathbf{g}_{m}-\mathbf{t}_{j}\right\|^{2} / \tau})}
\end{align}
where $\tau$ is the temperature hyperparameter. 

The categorical likelihood for cell $i$ and gene $m$ is defined as in the original ETM:
\begin{equation}  
    p(x_{im} | \boldsymbol{\theta}_i, \mathbf{B})
    = \prod^{n_i}_{\ell=1} r_{im}^{[\ell=m]} = r_{im}^{x_{im}}
\end{equation}
where $n_i$ is total number of transcripts in cell $i$ (e.g., Unique Molecular Identifier or UMI counts) and the categorical rate or the expected transcriptional rate for gene $m$ in cell $i$ is defined as:
\begin{equation}  
    r_{im} = \frac{\exp(\boldsymbol{\theta}_i\mathbf{b}^\top_m)}{\sum_m\exp(\boldsymbol{\theta}_i\mathbf{b}^\top_m)}
\end{equation}

\subsubsection{Overall loss function:}
The VAE loss function is defined as
\begin{align}\label{eq:l-vae}
    \mathcal{L}_{\mathrm{VAE}} & =\frac{1}{n} \sum_{i=1}^{n}-\mathbf{x}_i^{\top} \log \left(\mathrm{Softmax}\left( \boldsymbol{\theta}_i\cdot\mathbf{B}^{\top}\right)\right) + \mathrm{KL}\left[q\left(\boldsymbol{\theta}_i \mid \mathbf{x}_i\right) \| p\left(\boldsymbol{\theta}_i\right)\right]
\end{align}
The first term is the reconstruction error or negative log likliheood. The second term is the Kullback-Leibler (KL) divergence between the variational distribution and the standard Gaussian prior distribution:
\begin{align}	
		\mathrm{KL}[q(z|x)|| p(z)] 
		&=\E_{q(z|x)}\left[\log q(z|x)\right] - \E_{q(z)}\left[\log p(z)\right]\\
		&=\sum_k\left[-\frac{1}{2}\log \{\sigma_k^*\}^2 - \frac{1}{2} - \frac{1}{2} \log (2\pi) \right] - 
		\left[-\frac{1}{2} \log (2\pi) - \frac{1}{2} \{\mu_k^*\}^2- \frac{1}{2} \{\sigma_k^*\}^2\right]\\
		&=-\frac{1}{2}\left[\log \{\sigma_k^*\}^2 - \{\mu_k^*\}^2 -\{\sigma_k^*\}^2+1\right]
	\end{align}

Throughout the training process, cell embedding learning and topic modeling are co-optimized. We define the overall loss function of scE$^2$TM as the combination of $\mathcal{L}_{\mathrm{VAE}}$ (Eq. (\ref{eq:l-vae})), $\mathcal{L}_{\mathrm{CVE}}$ (Eq. (\ref{eq:l-cme})), and $\mathcal{L}_{\mathrm{ECR}}$ (Eq. (\ref{eq:l-ecr})), that is,
\begin{equation}
    \mathcal{L}=\mathcal{L}_{\mathrm{VAE}} +\mathcal{L}_{\mathrm{CVE}}+\lambda\cdot\mathcal{L}_{\mathrm{ECR}},
\end{equation}
where $\lambda$ is the weight hyperparameter. Algorithm \ref{alg:model} outlines the training workflow of scE$^2$TM.

\begin{algorithm}[htbp!]
\caption{Training algorithm for scE$^2$TM}\label{alg:model}
\textbf{Input:} observed single-cell gene expression $\mathbf{X}=\{\mathbf{x}_i\}$; cell embeddings from scGPT $\mathbf{V}=\{\mathbf{v}_i\}$.\\
\textbf{Hyperparameter:} number of epochs $n_e$; weight of the ECR loss $\lambda$.\\
\textbf{Output:} topic embeddings $\mathbf{T}=\{\mathbf{t}_k\}$; gene embeddings $\mathbf{G}=\{\mathbf{g}_m\}$; cell-topic matrix $\Theta=\{\boldsymbol{\theta}_i\}$; gene-topic distributions $\mathbf{B}$.
\begin{algorithmic}[1]
\State \textbf{for} $1$ \textbf{to} $n_e$ \textbf{do}
    \State \hspace{0.25cm} \textbf{for} each mini-batch from $\mathbf{X}$ and $\mathbf{V}$ \textbf{do}
    \State \hspace{0.5cm} $\boldsymbol{\theta}_i,\boldsymbol{\varphi}_i\leftarrow \mathrm{Encode}(\mathbf{x}_i,\mathbf{v}_i)$~\textcolor{blue}{$\triangleright$ Cross-view encoder}
    \State \hspace{0.5cm} Compute $\mathbf{B}$ by Eq. \eqref{eq:gene-topic}
    \State \hspace{0.5cm} $\hat{\mathbf{x}}_i\leftarrow \mathrm{Decode}(\boldsymbol{\theta}_i,\mathbf{B})$~\textcolor{blue}{$\triangleright$ Sparse linear decoder}
    \State \hspace{0.5cm} Compute $\mathcal{L}_{\mathrm{VAE}}$ by Eq. \eqref{eq:l-vae}
    \State \hspace{0.5cm}~\textcolor{blue}{$\triangleright$ CVE}
    \State \hspace{0.5cm} Sample neighbors $\mathbf{x}_{i}^{\mathcal{N}}\in\mathcal{N}\left(\mathbf{x}_{i}\right)$ and $\mathbf{v}_{i}^{\mathcal{N}}\in\mathcal{N}\left(\mathbf{v}_{i}\right)$
    \State \hspace{0.5cm} $\boldsymbol{\theta}^\mathcal{N}_i,\boldsymbol{\varphi}^\mathcal{N}_i\leftarrow \mathrm{Encode}(\mathbf{x}^\mathcal{N}_i,\mathbf{v}^\mathcal{N}_i)$~\textcolor{blue}{$\triangleright$ Cross-view encoder}
    \State \hspace{0.5cm} Compute $\mathcal{L}_{\mathrm{CVE}}$ by Eq. \eqref{eq:l-cme}
    \State \hspace{0.5cm}~\textcolor{blue}{$\triangleright$ ECR}
    \State \hspace{0.5cm} $C_{mk}=\| \mathbf{g}_m - \mathbf{t}_k \|^2,\forall m,k$
    \State \hspace{0.5cm} $\boldsymbol{\pi}_{\epsilon}^* \leftarrow \mathrm{Sinkhorn}(\mathbf{C})$
    \State \hspace{0.5cm} Compute $\mathcal{L}_{\mathrm{ECR}}$ by Eq. \ref{eq:l-ecr}
    \State \hspace{0.5cm}~\textcolor{blue}{$\triangleright$ Parameter Inference}
    \State \hspace{0.5cm} Compute $\mathcal{L}=\mathcal{L}_{\mathrm{VAE}} +\mathcal{L}_{\mathrm{CVE}}+\lambda\cdot\mathcal{L}_{\mathrm{ECR}}$
    \State \hspace{0.5cm} Update $\mathbf{T}$, $\mathbf{G}$, $\Theta$, $\mathbf{B}$ with a gradient step
\end{algorithmic}
\end{algorithm}




\subsubsection{Software implementation}
Our scE$^2$TM is implemented in Python and the core model is based on the PyTorch (v.1.9.8) framework. Throughout the training process, we set the dimension size of each hidden layer to 200 and used $\tanh()$ as the activation function. The optimization of scE$^2$TM is performed using the RMSprop optimizer with 500 iterations. For datasets with more than 10,000 cells, we uniformly use a batch size of 2048, otherwise it is set to 512.
For $\lambda$, following the setup of  Wu \textit{et al} \cite{wu2023effective}, its value is set to 20 or 100 for different datasets. The weight $\alpha$ of $\mathcal{L}_{\mathrm{REG}}$ is set to 5.
Furthermore, for fair comparison, we followed previous work \cite{zhao2021learning,chen2024comprehensive} in setting the number of topics $K$ to 100 for all single-cell embedded topic models. When calculating TC, TD, TQ, ORA$_N$, ORA$_U$, and ORA$_Q$, we set the number of top genes $h$ for each topic to 10. All experiments, except for the embedding extraction of the foundation model, were conducted in a Python environment equipped with an Nvidia RTX 3090 GPU and 128G RAM. 

\subsection{Differential gene expression and topic analysis}
\label{sec:topicdifferential}
For differential gene expression analysis, we compute the log fold-change in expression between
cells with and without a given label. Statistical significance is assessed using the Mann-Whitney U test, as implemented in ``\texttt{scanpy.tl.rank\_genes\_groups()}''.
$P$-values are derived from the U statistic using a normal approximation with tie correction, and multiple testing correction is performed using the BH method.

For differential topic analysis, we aim to identify topics significantly associated with specific cell types or disease states. For each topic-cell label pair, we compute the difference in mean topic activity between cells with and without the label. 
Statistical significance is evaluated using the same Mann-Whitney U test framework described above, followed by BH correction for multiple comparisons. 
A topic is considered specific to a cell type or condition if its BH-adjusted $q$-value is below 0.01 and the mean difference for cell type-related topics exceeds 1. We use the CellMarker \cite{hu2023cellmarker} database to examine the overlap between top genes in cell type-specific differentially expressed topics and known cell type markers.

\subsection{Topic-based \textit{in-silico} perturbations experiments}
\label{sec:perturbation}
In the IFN-$\beta$ PBMC perturbation experiment, we increased the treatment-specific topics among untreated cells and observe their changes in term of the expected gene expression output from the decoder. For each control cell, we set the topic scores to 0.1, 0.15, and 0.2.
To study perturbations effects of the malignant-specific topics, we decrease the topic scores in the malignant cells by setting them to 1/8, 1/16, and 0 of their original values. This simulated varying degrees of malignant topic suppression. As negative control experiments for both case studies, we applied two types of controls. First, we randomly selected the same number of non-condition-specific topics and repeated the analysis (random topics). Second, we selected an equal number of topics that showed the lowest correlation with the condition (non-malignant or non-IFN specific topics).


\subsection{Survival analysis}
We extracted clinical and scRNA-seq data of melanoma patients from the TCGA database using the \texttt{cgdsr} R software package. Following an existing protocol \cite{chen2024comprehensive}, patient
samples were divided into two groups corresponding to low (25\%) and high (75\%) expression of
the target gene. We ran Kaplan-Meier algorithm implemented in the survival R package to estimate the survival functions of the two groups. The same analysis was performed for the malignant-specific topics.

\subsection{Categorization of perturbed cells}\label{sec:svm}
We divided the data into training and testing sets in a 7:3 ratio, which were used to train and test a binary SVM classifier to predict treatment or condition label of a cell. The reported results represent the average of 10 independent runs.

\subsection{Code availability}
The scE$^2$TM model and all the code necessary for reproducing our results is publicly available via GitHub at \url{https://github.com/nbnbhwyy/scE2TM}. To ensure reproducible results, all analyses are performed using the same fixed random seed. An archived version will be deposited in the Zenodo database upon acceptance.

\section{Acknowledgments}
Y.L. is supported by Canada Research Chair (Tier 2) in Machine Learning for Genomics and Healthcare (CRC-2021-00547) and CIHR Project Grant (202503PJT-540722-GMX-CFAA-255237). Y.R. is supported by the National Natural Science
Foundation of China (62372483).

\section{Competing interests}
The authors declare that they have no competing interests.

\bibliography{sn-bibliography}
 
\end{document}